\documentclass[sigconf, screen]{acmart}
\AtBeginDocument{%
  }

\copyrightyear{2026}
\acmYear{2026}
\setcopyright{cc}
\setcctype{by}
\acmConference[CSLAW '26]{Symposium on Computer Science and Law}{March 03--05, 2026}{Berkeley, CA, USA}
\acmBooktitle{Symposium on Computer Science and Law (CSLAW '26), March 03--05, 2026, Berkeley, CA, USA}
\acmPrice{}
\acmDOI{10.1145/3788646.3789522}
\acmISBN{979-8-4007-2447-3/2026/03}

\usepackage[most]{tcolorbox}
\tcbuselibrary{most}
\usepackage{wrapfig}
\usepackage{float}
\usepackage{tabularx}

\definecolor{cyan10}{HTML}{E5F6FF}
\definecolor{cyan20}{HTML}{BAE6FF}
\definecolor{cyan60}{HTML}{0072c3}
\definecolor{cyan70}{HTML}{00539a}
\definecolor{cyan80}{HTML}{003a6d}
\definecolor{teal10}{HTML}{D9FBFB}
\definecolor{teal20}{HTML}{9EF0F0}
\definecolor{teal60}{HTML}{007d79}
\definecolor{orange10}{HTML}{FFF2E8}
\definecolor{orange20}{HTML}{FFD9BE}
\definecolor{orange60}{HTML}{ba4e00}
\definecolor{blue10}{HTML}{EDF5FF}
\definecolor{blue20}{HTML}{D0E2FF}
\definecolor{blue70}{HTML}{0043ce}
\definecolor{blue80}{HTML}{002d9c}
\definecolor{magenta10}{HTML}{FFF0F7}
\definecolor{magenta20}{HTML}{FFD6E8}
\definecolor{magenta30}{HTML}{ffafd2}
\definecolor{magenta50}{HTML}{ee5396}
\definecolor{magenta60}{HTML}{d02670}
\definecolor{magenta70}{HTML}{9f1853}
\definecolor{purple10}{HTML}{F6F2FF}
\definecolor{purple20}{HTML}{E8DAFF}
\definecolor{purple30}{HTML}{d4bbff}
\definecolor{purple70}{HTML}{8a3ffc}
\definecolor{rose10}{HTML}{FCF2ED}
\definecolor{rose20}{HTML}{F9D9D1}
\definecolor{rose60}{HTML}{ab5638}
\definecolor{rose70}{HTML}{853c27}
\definecolor{red10}{HTML}{FFF1F1}
\definecolor{red20}{HTML}{FFD7D9}
\definecolor{green10}{HTML}{DEFBE6}
\definecolor{green20}{HTML}{A7F0BA}
\definecolor{green70}{HTML}{0e6027}
\definecolor{green80}{HTML}{044317}
\definecolor{yellow10}{HTML}{fcf4d6}
\definecolor{yellow20}{HTML}{fddc69}
\definecolor{gray20}{HTML}{e0e0e0}
\definecolor{gray30}{HTML}{c6c6c6}
\definecolor{gray40}{HTML}{a8a8a8}
\definecolor{gray80}{HTML}{393939}
\definecolor{red60}{HTML}{da1e28}
\definecolor{green60}{HTML}{00a78f}

\begin{document}

\title{Thinking Longer, Not Always Smarter: Evaluating LLM Capabilities in Hierarchical Legal Reasoning}

\author{Li Zhang}
\authornote{Corresponding author}
\email{liz239@pitt.edu}
\orcid{0000-0003-0375-1793}
\affiliation{%
  \institution{University of Pittsburgh}
  \city{Pittsburgh}
  \state{Pennsylvania}
  \country{USA}
}

\author{Matthias Grabmair}
\email{matthias.grabmair@tum.de}
\orcid{0000-0001-6586-2486}
\affiliation{%
  \institution{Technical University of Munich}
  \city{Munich}
  \state{Bavaria}
  \country{Germany}
}

\author{Morgan Gray}
\email{mag454@pitt.edu}
\orcid{0000-0002-3800-2103}
\affiliation{%
  \institution{University of Pittsburgh}
  \city{Pittsburgh}
  \state{Pennsylvania}
  \country{USA}
}

\author{Kevin Ashley}
\email{ashley@pitt.edu}
\orcid{0000-0002-5535-0759}
\affiliation{%
  \institution{University of Pittsburgh}
  \city{Pittsburgh}
  \state{Pennsylvania}
  \country{USA}
}

\renewcommand{\shortauthors}{Li et al.}

\begin{abstract}
  Case-based reasoning is a cornerstone of U.S. legal practice, requiring professionals to argue about a current case by drawing analogies to and distinguishing from past precedents. While Large Language Models (LLMs) have shown remarkable capabilities, their proficiency in this complex, nuanced form of reasoning needs further investigation. We propose a formal framework that decomposes the process of identifying significant distinctions between cases into three-stage reasoning tasks. Our framework models cases using factual predicates called factors, organizes them into a legal knowledge hierarchy, and defines verifiable rules for identifying distinctions, analyzing their argumentative support, and evaluating their significance. Through comprehensive evaluation of modern reasoning LLMs, we reveal a paradox: while models achieve high accuracy on surface-level reasoning (Task 1), performance degrades on hierarchical reasoning (Task 2: 64.82\%-92.09\%) and collapses on integrated analysis (Task 3: 11.46\%-33.99\%). Most strikingly, we find that models consistently expend more computational resources on incorrect responses than correct ones, suggesting that ``thinking longer'' does not always mean ``thinking smarter.'' Our work provides a methodology for fine-grained analysis of LLM reasoning capabilities in complex domains and reveals fundamental limitations that must be addressed for robust and trustworthy legal AI. Our project page is available at \url{https://thinking-longer-thinking-smarter.vercel.app}.
\end{abstract}

\ccsdesc[500]{Computing methodologies~Natural language processing}
\ccsdesc[500]{Applied computing~Law}
\keywords{Large Language Model, Hierarchical Reasoning, Case-based Reasoning, Trustworthy AI, Evaluation}


\maketitle

\section{Introduction}

\begin{figure*}[t]
  \centering
  \includegraphics[width=0.85\textwidth]{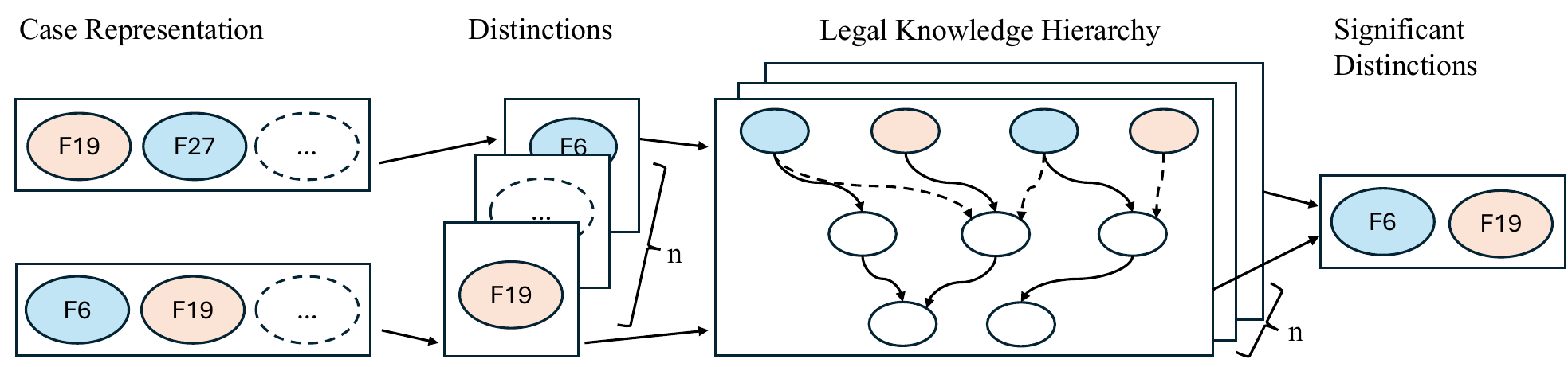}
  \caption{The decomposed framework for identifying significant distinctions, which consists of three steps: (1) identify distinctions, (2) analyze argumentative roles of a distinction via legal knowledge hierarchy, and (3) identify significant distinctions. Red and blue presents the favoring side of the factors.}
  \label{fig:teaser}
  \end{figure*}

A core component of legal expertise is the ability to reason from precedent~\cite{lamond2006precedent, ashley2017artificial}. In the common law system, the doctrine of stare decisis obligates courts to follow the rulings of prior cases with similar factual circumstances~\cite{douglas1949stare}. Attorneys, therefore, build arguments by drawing analogies to favorable precedents and distinguishing their current case from unfavorable ones~\cite{gerhardt2011power}. This process is not a simple pattern-matching task~\cite{schauer2017analogy, horty2011rules, schauer2008precedent}; its reasoning process involves abstracting specific facts into broader legal concepts, weighing conflicting evidence, and understanding how different facts contribute to an overall legal argument.

The recent success of LLMs has raised the question of whether they can assist in these sophisticated reasoning tasks~\cite{smith2025history, atkinson2019reasoning}. Reasoning LLMs have shown remarkable capabilities in various domains~\cite{ke2025survey}, including planning~\cite{wei2025plangenllms} and mathematical reasoning~\cite{wang2025survey}. However, their capabilities in the complex, nuanced form of reasoning required in legal practice needs further investigation. This gap is particularly concerning given the increasing interest in deploying AI systems in legal contexts, where accuracy and reliability are paramount.

We introduce a formal framework that deconstructs one of the central arguments in case-based reasoning: identifying significant distinctions. A significant distinction is a factual difference between a current case and a precedent that is critical enough to justify a different outcome. Our framework operationalizes this concept through a sequence of three increasingly complex reasoning tasks, each designed to integrate specific aspects of hierarchical legal reasoning.

We build a decomposed framework grounded in the CATO tradition of case-based legal reasoning~\cite{aleven1997teaching, ashley1997reasoning, aleven1995doing}. Each problem includes a factor hierarchy, a current case ($C_1$), and a precedent case ($C_2$). A symbolic solver computes ground truth using formal rules about distinctions, support, blocking, and emphasis/downplay. We then prompt LLMs to perform the same analysis in three increasingly challenging tasks. Our overall framework, shown in Figure~\ref{fig:teaser}, provides a comprehensive view of how cases are processed from their linear representations to the identification of significant distinctions through hierarchical legal reasoning.

Our evaluation reveals a paradox that challenges conventional assumptions about LLM reasoning. While models achieve high accuracy on surface-level reasoning (Task 1), performance degrades significantly on hierarchical reasoning (Task 2: 64.82\%-92.09\%) and collapses on integrated analysis (Task 3: 11.46\%-33.99\%). Most remarkably, we find that models consistently expend more computational resources on incorrect responses than correct ones, suggesting that ``thinking longer'' does not always mean ``thinking smarter.'' This finding has profound implications for the development of reasoning LLMs.

Our contributions are threefold. First, we operationalize the CATO-style hierarchy into three decomposed tasks: (1) identify distinctions, (2) assess emphasizing vs. downplaying specific distinctions, and (3) integrate all steps to identify all significant distinctions. Second, we formalize the process of identifying significant distinctions using a symbolic representation of cases, factors, and a conceptual hierarchy. This model bridges abstract legal theory and computational implementation. Third, we present an empirical study that reveals limitations in current LLM architectures for complex legal reasoning, providing insights for the development of more effective and trustworthy legal AI systems.

\section{Related Work}


\subsection{AI \& Law: Case-Based Reasoning Systems}

Early AI and law research established the foundation for computational legal reasoning. The HYPO system~\cite{ashley1991reasoning, ashley1992case} pioneered factor-based case representation and distinguishing cases through factual differences and explored hypothetical reasoning. The CATO system~\cite{aleven1997teaching, ashley1997reasoning} introduced hierarchical organization of legal knowledge, recognizing that factors serve as evidence for abstract legal concerns, which support higher-level legal issues. CATO's framework for identifying distinctions, analyzing their argumentative roles, and determining significance provided the theoretical foundation for our decomposed evaluation approach.

Subsequent work extended these systems through CABARET~\cite{rissland1991cabaret}, which integrated case-based with rule-based reasoning,  IBP~\cite{bruninghaus2003predicting} which employed machine learning to identify factors and case-based argumentation to predict outcomes, and VJAP~\cite{grabmair2017predicting, grabmair2016modeling}, which developed a value-based legal argumentation system. Recent research has focused on scaling these approaches to larger case databases and more complex legal domains~\cite{chau2024computational,zhang2025mitigating, gray2025generating}.

These traditional systems required extensive manual knowledge engineering and operated on structured representations with explicit rule-based reasoning. The emergence of LLMs has created new opportunities for legal AI, but also new challenges in evaluating and understanding  their reasoning capabilities. Our work bridges this gap by providing a framework for evaluating how well modern LLMs can perform the sophisticated reasoning tasks that these early systems were designed to handle.

\subsection{Hierarchical Reasoning in LLMs}

Hierarchical reasoning is crucial for complex reasoning tasks like legal analysis~\cite{wang2025hierarchical, lyu2023multi, canavotto2023reasoning}. It involves moving between different levels of abstraction and integrating information across multiple levels. For example, in legal case analysis, a lawyer needs to reason from specific facts to intermediate concepts and ultimately to high-level legal issues.

Chain-of-thought prompting~\cite{wei2022chain} demonstrated that explicitly encouraging models to generate intermediate reasoning steps can significantly improve performance on complex tasks. However, chain-of-thought approaches typically focus on linear reasoning chains rather than the hierarchical structures found in legal reasoning~\cite{wang2025hierarchical}. Tree-of-thoughts approaches~\cite{yao2023tree} have extended this work to explore multiple reasoning paths simultaneously, allowing models to consider alternative approaches and backtrack when necessary. Recent work on reasoning with external tools~\cite{mialon2023augmented, ke2025survey} has explored how models can leverage specialized reasoning modules for different types of analysis.

Most existing work on hierarchical reasoning in LLMs has focused on mathematical or logical reasoning tasks. The legal domain presents unique challenges due to the need to reason about abstract legal concepts, weigh conflicting evidence, and integrate multiple types of analysis. Our work addresses this gap by providing a specific framework for evaluating hierarchical reasoning in the legal domain.

\subsection{Alignment and Evaluation Methodologies for LLMs}

The deployment of AI systems in legal contexts raises questions about trustworthiness, reliability, and accountability~\cite{afroogh2024trust, kattnig2024assessing}. Legal professionals need to understand not just what AI systems can do, but how they reason and where they may fail~\cite{zhang2025measuring, zhang2025mitigating}.

The evaluation of AI systems in legal contexts presents unique challenges due to the complexity and nuance of legal reasoning~\cite{posner2025judge}. Traditional evaluation approaches~\cite{aletras2016predicting, chalkidis2018obligation, kim2017two} focus on accuracy on specific tasks like judicial decision prediction, element extraction and question answering, which may not capture the full range of capabilities required for legal practice~\cite{kattnig2024assessing}. Recent work has developed benchmarks such as CUAD~\cite{hendrycks2021cuad}, LegalBench~\cite{guha2023legalbench} and the COLIEE competition~\cite{goebel2025international}, but these focus primarily on isolated skills rather than the integrated application of the reasoning chain.

Recent work on explainable AI has focused on providing interpretable explanations for model decisions~\cite{bilal2025llms}, but most work has focused on tasks like text classification~\cite{chhatwal2018explainable, collenette2023explainable} and judgement prediction~\cite{collenette2020explainable, sesodia2025annocaselaw} and may not be applicable to the reasoning required in legal analysis. Work on AI safety and alignment~\cite{irving2018ai, buhl2025alignment, hubinger2024sleeper} has explored ensuring AI systems behave consistently with human values and expectations, which is particularly relevant to legal AI where systems must reason consistently with legal principles and professional standards.

Our work addresses these challenges by providing a decomposed evaluation framework on legal reasoning. We structure the evaluation around a principled reasoning process instead of atomic components, enabling detailed analysis of where models succeed and fail, and guiding the development of more capable and trustworthy legal AI systems.

\section{Modeling Hierarchical Case-Based Legal Reasoning}

To computationally model case-based reasoning, we first formalize its core components: the cases themselves and the hierarchical structure of legal concepts they relate to.

\subsection{Case Representation}

A legal case can be distilled into a set of legally relevant facts, represented as factors, and an outcome. Factors are stereotyped fact patterns that tend to favor either the plaintiff (p) or the defendant (d).

\textbf{Definition 1 (Factor)}: A factor $f$ is a predicate representing a legally relevant fact. The set of all possible factors is denoted by $\mathcal{F}$. Each factor $f \in \mathcal{F}$  has a favored party, $s(f) \in \{p, d\}$.

\textbf{Definition 2 (Case)}: A case $C$ is a pair $F, o$, where $F \subseteq \mathcal{F}$ is the set of factors present in the case, and $o \in \{p, d\}$ is the outcome (i.e., the winning party).

In our analysis, we consider a current case, $C_1 = (F_1, o_1)$, where the outcome $o_1$ is unknown, and a precedent case, $C_2 = (F_2, o_2)$, where the outcome $o_2$  is known.

\subsection{Hierarchy of Legal Knowledge}

Factors do not exist in isolation. They serve as evidence for more abstract legal concerns, which in turn inform the ultimate legal issues in a case. This multi-level abstraction is central to legal argument, allowing legal professionals to argue that, while the surface-level facts may differ, the underlying legal principles might be the same. We model this as a hierarchy.

\textbf{Definition 3 (Hierarchy and Nodes)}: The hierarchy is a directed acyclic graph (DAG) $G = (V, E)$ where:

- $V = \mathcal{F} \cup \mathcal{C} \cup \mathcal{I}$ is the set of nodes, comprising base-level factors $\mathcal{F}$, intermediate legal concerns $\mathcal{C}$, and top-level legal issues $\mathcal{I}$.

- $E$ is the set of edges, representing support relationships. An edge from node $u$ to $v$ means $u$ provides evidence for $v$.

Not all support is equal. For instance, a factor can provide strong, direct evidence for a concern, or weak, circumstantial evidence.

\textbf{Definition 4 (Edge Strength)}: Each edge $e \in E$ has a strength, $\sigma(e) \in \{\text{strong}, \text{weak}\}$.

This hierarchy forms the backbone of our reasoning framework, allowing us to model the nuanced connections between concrete facts and abstract legal concepts.

\section{A Decomposed Framework for Identifying Significant Distinctions}
\label{sec:framework}

\begin{figure}[t]
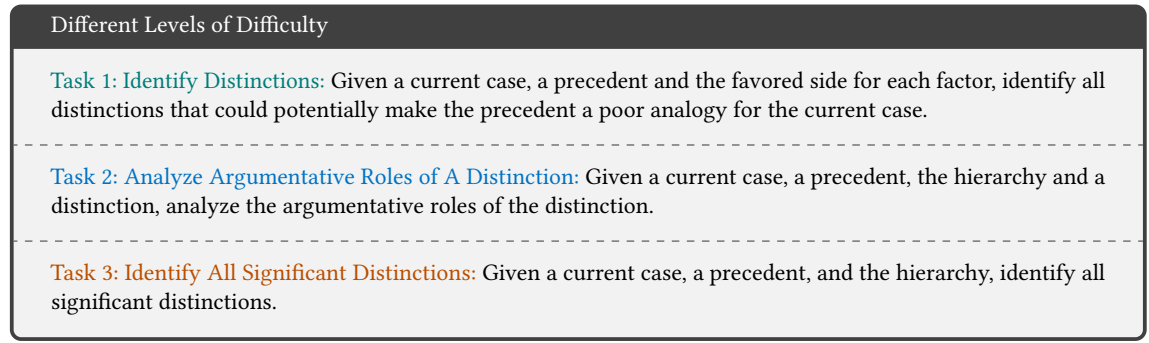

  \centering
  \begin{tcolorbox}[
    enhanced,
    title=Different Levels of Difficulty,
    separator sign={\tcbline},
    separator sign dash={3pt}{3pt},
  ]
  \textcolor{teal60}{Task 1: Identify Distinctions:} Given a current case, a precedent and the favored side for each factor, identify all distinctions that could potentially make the precedent a poor analogy for the current case.
  \tcbline
  \textcolor{cyan60}{Task 2: Analyze Argumentative Roles of A Distinction:} Given a current case, a precedent, the hierarchy and a distinction, analyze the argumentative roles of the distinction.
  \tcbline
  \textcolor{orange60}{Task 3: Identify All Significant Distinctions:} Given a current case, a precedent, and the hierarchy, identify all significant distinctions.
  \end{tcolorbox}
  \vspace{-2mm}
  \caption{Modifying the difficulty level of the tasks by increasing the reasoning complexity.}
  \label{fig:difficulty_examples}
\end{figure}

We now present our core contribution: a decomposed framework that formalizes the identification of significant distinctions. As illustrated in Figure \ref{fig:teaser}, our framework transforms the complex process of hierarchical legal reasoning from linear case representations through the systematic extraction of distinctions, incorporating legal knowledge hierarchies and formal rules to ultimately identify significant distinctions. The framework consists of three sequential tasks, each with increasing complexity as demonstrated in Figure \ref{fig:difficulty_examples}. A step-by-step application of this framework is detailed in Appendix~\ref{sec:illustrative_example}.

\subsection{Task 1: Identify Distinctions}

The first step is to identify all factual differences that could potentially make the precedent a poor analogy for the current case. These are called distinctions.

\textbf{Definition 5 (Distinction)}: Given a current case $C_1=(F_1, o_1)$ and a precedent $C_2=(F_2, o_2)$, a factor $f \in \mathcal{F}$ is a distinction if it satisfies one of the following conditions:

\begin{enumerate}
\item The factor is a strength for the precedent's winner that the current case lacks:
$f \in F_2 \land f \notin F_1 \land s(f) = o_2$
\item The factor is a weakness for the precedent's loser that is present in the current case:
$f \in F_1 \land f \notin F_2 \land s(f) \neq o_2$
\end{enumerate}

\textbf{Input}: A current case $C_1=(F_1, o_1)$, a precedent $C_2=(F_2, o_2)$ and the favored side $s(f)$ for each factor $f \in F_1 \cup F_2$.

\textbf{Goal}: To produce the set of all distinctions, $D = \{f \in F_1 \cup F_2 \mid f \text{ is a distinction}\}$.

\textbf{Metrics}:
    \begin{itemize}
        \item Accuracy: For a model-generated set of distinctions $D_{model}$, accuracy is 1 if $D_{model} = D_{gt}$  (exact set match with the ground truth set), and 0 otherwise. Overall accuracy is the average over all test instances.
        \item Reasoning Token: The average reasoning token cost for the model to generate a response for a single instance.
    \end{itemize}

\subsection{Task 2: Analyzing Argumentative Roles of A Distinction}

A factor's effectiveness depends on how it supports the legal concerns in the hierarchy. This support can be direct and strong, or it can be weak and potentially neutralized by countervailing evidence. This leads to the concepts of ``effective support'' and ``blocking.''

\textbf{Definition 6 (Path)}: A path $\pi(f, P)$  from a factor $f$ to an ancestor concern $P$ is a sequence of edges in the hierarchy. The path is strong if all edges in it are strong. If any edge is weak, the entire path is weak.

\textbf{Definition 7 (Blocking)}: In a case $C_i$, the support from a factor $(f \in F_i)$  for a concern $P$ is blocked if and only if:
\begin{enumerate}
\item The path $\pi(f, P)$ is weak.
\item There exists an opposing factor $f_{\text{con}} \in F_i$ (where $s(f_{\text{con}}) \neq s(f)$) that has a strong path to $P$.
\end{enumerate}

\textbf{Definition 8 (Effective Support)}: A factor $f$ provides effective support for a concern $P$ if its path to $P$ is strong, or if its path is weak and not blocked.

This task assesses a model's ability to trace dependencies in a hierarchy and apply a rule-based system for weighing conflicting evidence, which simulates the real-world case-based legal reasoning.

Once a distinction is identified, its argumentative power can be modulated. An opponent may downplay it by pointing to other facts that serve the same argumentative purpose. Conversely, one can emphasize it by showing it creates a crucial argumentative gap.

Let $D$ be a distinction. Assume $D$ is a pro-plaintiff factor present in the precedent $C_2$ but not the current case $C_1$. Let $P$ be an ancestor concern of $D$.

\textbf{Definition 9 (Downplaying a Distinction)}: The distinction $D$ can be downplayed with respect to $P$  if, in the current case $C_1$, there exists an alternative pro-plaintiff factor $f_{\text{alt}} \in F_1$ $(f_{\text{alt}} \neq D)$ that provides effective support for $P$.

\textbf{Definition 10 (Emphasizing a Distinction)}: The distinction $D$ can be emphasized with respect to $P$ if the current case $C_1$ lacks any pro-plaintiff factor that provides effective support for $P$.

This task requires reasoning about hypotheticals and alternative arguments, a sophisticated cognitive skill.

\textbf{Input}: A current case $C_1$, a precedent $C_2,$ the hierarchy $G$, and a specific distinction $D$ to analyze.

\textbf{Goal}: To produce a boolean pair, $(\text{can\_be\_emphasized}, \newline \text{can\_be\_downplayed})$, indicating if $D$ can be emphasized and if it can be downplayed with respect to any of its ancestor concerns, respectively.

\textbf{Metrics}:
    \begin{itemize}
        \item Accuracy: An instance is correct only if both boolean values match the ground truth pair.
        \item Reasoning Token: The average reasoning token cost for the model to generate a response for a single instance.
    \end{itemize}

\subsection{Task 3: Identify Significant Distinctions}

The final step synthesizes the previous analyses to determine all significant distinctions.

\textbf{Definition 11 (Significant Distinction)}: A distinction $D$ is significant if and only if it can be emphasized and it cannot be downplayed.

A significant distinction represents a fundamental difference between the cases that is not easily explained away. It points to a core weakness in the analogical argument and provides a strong basis for arguing for a different outcome.

\textbf{Input}: A current case $C_1=(F_1, o_1)$, a precedent $C_2=(F_2, o_2)$, and the hierarchy $G$.

\textbf{Goal}: To produce the set of all significant distinctions, $D^{sig} = \{f \in F_1 \cup F_2 \mid f \text{ is a significant distinction}\}$.

\textbf{Metrics}:
    \begin{itemize}
        \item Accuracy: For a LLM-generated set $D_{model}^{sig}$, accuracy is 1 if $D_{model}^{sig} = D_{gt}^{sig}$ (exact set match with the ground truth set), and 0 otherwise.
        \item Reasoning Token: The average reasoning token cost for the model to generate a response for a single instance.
    \end{itemize}

\section{Experimental Setup}

\begin{figure*}[t]
  \centering
  \includegraphics[width=0.85\textwidth]{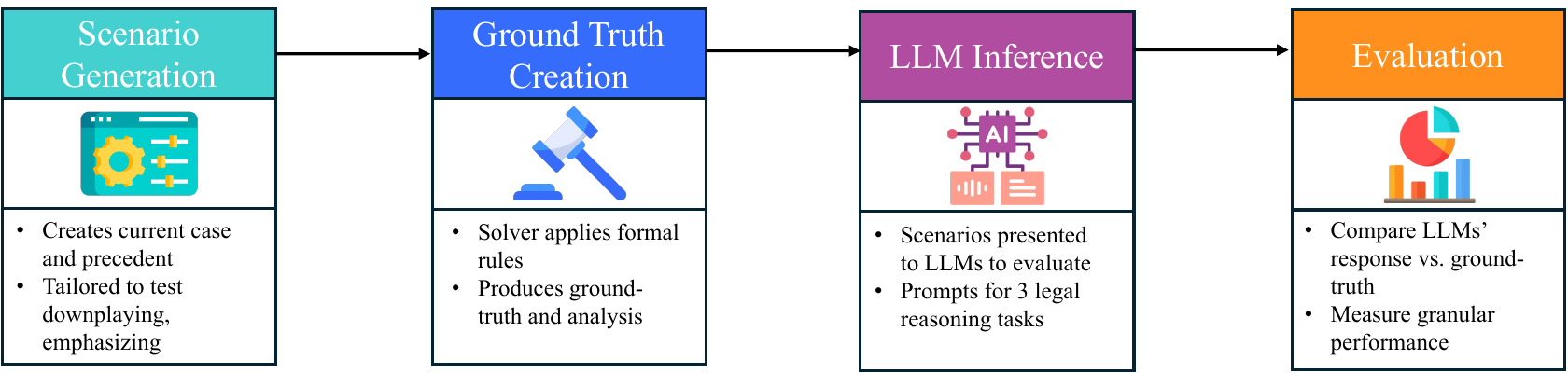}
  \caption{The evaluation pipeline, which consists of scenario generation, ground truth creation, LLM inference, and evaluation.}
  \label{fig:pipeline}
\end{figure*}

\subsection{Evaluation Pipeline}

To evaluate LLMs against our framework, we implemented an evaluation pipeline that operationalizes the three tasks described in Section \ref{sec:framework}. This pipeline as shown in Figure \ref{fig:pipeline} allows us to test models in a controlled and reproducible manner.

\begin{itemize}
    \item Scenario Generation: We developed a scenario generator that creates pairs of a current case and a precedent in factor-based representation by sampling factors from the CATO framework (see Tables \ref{tab:factormr}, \ref{tab:issues}, and \ref{tab:concerns} for the list of factors, issues, and concerns). The generator is designed to produce scenarios that are tailored to test each of the three tasks, creating cases with specific instances of blocking, opportunities for downplaying/emphasizing, etc.
    \item Ground Truth Creation: For each generated scenario, a deterministic solver applies the formal rules defined in Section \ref{sec:framework} to produce a ground-truth analysis for each task.
    \item LLM Inference: The scenarios are then presented in a structured format to LLMs, which are prompted to perform each of the three reasoning tasks. The detailed prompt templates used for each task are provided in Appendix \ref{sec:prompts}.
    \item Evaluation: The LLM-generated responses are compared against the ground truth to calculate accuracy for each task. This allows us to measure performance at a granular level.
\end{itemize}

\subsection{Case Complexity}

Our experimental setup employed calibrated parameters across different model families to ensure fair comparison while optimizing for each model's capabilities. The evaluation framework used 253 test instances on each task, where each case pair consisted of a current case with 4 factors and a precedent case with 4 factors, representing moderate complexity scenarios. The legal knowledge hierarchy was represented using the CATO framework in Mermaid diagram format (see Figure \ref{fig:cato}), providing the relationships between factors, legal concerns, and legal issues.

\subsection{Reasoning LLMs to Evaluate and Configuration}

We evaluated the following 5 reasoning LLMs: gpt-5-2025-08-07(gpt-5)~\cite{openai_gpt5_system_card}, qwen3-235b-a22b-thinking-2507(qwen3-thinking)~\cite{yang2025qwen3}, gpt-oss-120b(gpt-oss-120b)~\cite{agarwal2025gpt}, gemini-2.5-pro-preview-06-05(gemini-pro)~\cite{comanici2025gemini}, gemini-2.5-flash-preview-05-20(gemini-flash)~\cite{comanici2025gemini}. We also evaluated a non-reasoning LLM: qwen3-235b-a22b-instruct-2507(qwen3-non-thinking)~\cite{yang2025qwen3}.

\textbf{Model-Specific Parameter Configuration:}\footnote{For gpt-oss-120b, it is recommended to use \texttt{temperature}=1.0 and \texttt{top\_p}=1.0~\cite{unsloth_docs}. Qwen team recommends \texttt{temperature}=0.6 and \texttt{top\_p}=0.95 for the thinking model and \texttt{temperature}=0.7 and \texttt{top\_p}=0.8 for the non-thinkingmodel~\cite{qwen3-235b-thinking, qwen3-235b-non-thinking}. We set \texttt{temperature}=0.3 and \texttt{top\_p}=0.95 based on the models' performance on other reasoning benchmarks to maintain sufficient consistency for reliable evaluation. We avoided \texttt{temperature}=0 to prevent overly deterministic responses that might miss alternative reasoning paths.}

Gemini Models (gemini-pro, gemini-flash): We configured these models with \texttt{temperature} = 0.3 and \texttt{top\_p} = 0.95. The \texttt{max\_tokens} limit was set to 65,536.

OpenAI Models (gpt-5, gpt-oss-120b): For the gpt-5 model, we used the default parameters because those parameters are not exposed to the API users. For gpt-oss-120b, we explicitly set the reasoning effort to be ``medium'' in the prompt and the \texttt{temperature} = 0.3, \texttt{top\_p} = 0.95 and the \texttt{max\_tokens} limit was set to 65,536.

Qwen Models (qwen3-thinking, qwen3-non-thinking): The Qwen models were configured with \texttt{temperature} = 0.3 and \texttt{top\_p} = 0.95 and \texttt{max\_tokens} = 65,536.

\section{Empirical Results}

\begin{figure*}[ht]
\centering
\includegraphics[width=0.94\textwidth]{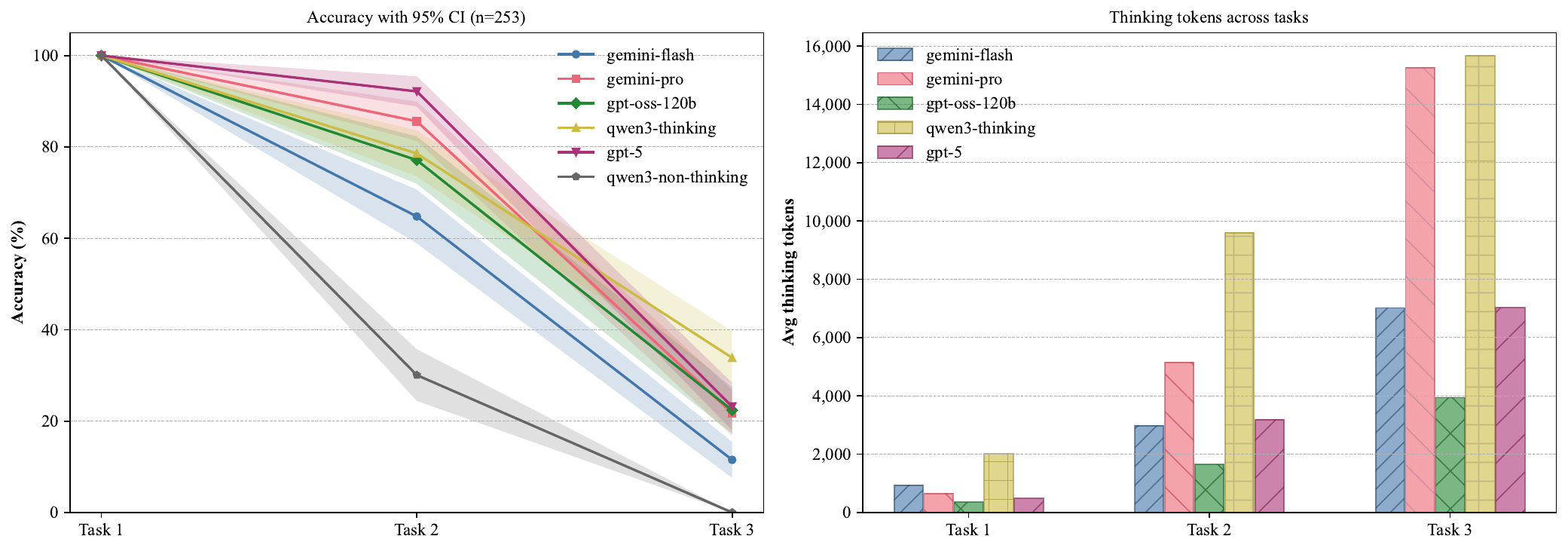}
\caption{Model performance across Tasks 1--3. The left panel illustrates accuracy, showing a decline as tasks become more complex. The right panel displays the average number of thinking tokens used, which increases with task difficulty.}
\label{fig:acc_tokens}
\end{figure*}

Our experimental evaluation, with results summarized in Figure \ref{fig:acc_tokens} and Table \ref{tab:accuracy-thinking-tokens}, reveals findings on the capabilities and limitations of current reasoning LLMs in hierarchical legal reasoning tasks.

\subsection{Performance Degradation Across Task Complexity}

We observe systematic decline in model accuracy as task complexity increases. All models achieve perfect performance (100\% accuracy) on Task 1, which requires identifying surface-level distinctions between cases. This suggests that the LLMs evaluated in this study excel at simple pattern recognition and basic comparative analysis when the reasoning requirements are relatively straightforward.

Performance degrades when models must engage in hierarchical reasoning about argumentative roles (Task 2), with accuracy ranging from 64.82\% (gemini-flash) to 92.09\% (gpt-5). The performance collapse occurs in Task 3, where models must synthesize all previous analyses to identify all significant distinctions. Accuracy of reasoning LLMs plummets to a range of 11.46\% (gemini-flash) to 33.99\% (qwen3-thinking), with most models achieving less than 25\% accuracy. This suggests that the integration of multiple reasoning steps represents a challenge for current reasoning LLMs.

\subsection{The Importance of Intermediate Reasoning}

The comparison between thinking and non-thinking models reveals the performance gain from the intermediate reasoning process. The qwen3-thinking model achieves 78.66\% accuracy on Task 2 compared to 30.04\% for its non-thinking counterpart, representing a 2.6-fold improvement. The advantage of thinking models becomes even more pronounced in Task 3, where the non-thinking model fails completely (0.00\% accuracy) while the thinking model maintains some capability (33.99\%). 

This difference demonstrates that reasoning capabilities enabled by the reinforcement-learning-based post-training strategy \cite{guo2025deepseek} are beneficial for navigating the hierarchical structure of legal knowledge. However, the diminishing returns observed in Task 3 indicate that current post-training methods on reasoning models have limitations in handling the complex multi-step legal analysis.

\subsection{Computational Efficiency and Reasoning Quality}

All models exhibit substantial increases in token usage as task complexity increases, with gpt-5 showing a 14.4-fold increase from Task 1 (487 tokens) to Task 3 (7,025 tokens). However, the correlation between computational effort and performance is not straightforward. qwen3-thinking uses the most tokens across all tasks (2,011 tokens for Task 1, 9,596 for Task 2, and 15,678 for Task 3) but does not achieve the highest accuracy in Task 2. In contrast, gpt-5 achieves the highest accuracy on Task 2 (92.09\%) while using significantly fewer tokens (3,189) than qwen3-thinking (9,596).

The performance patterns reveal distinct characteristics of different model families. Gpt-5 demonstrates the most efficient reasoning in this study, achieving high accuracy with relatively moderate token usage. Gemini-pro shows interesting efficiency characteristics, using fewer tokens than gemini-flash on Task 1 but achieving higher accuracy with more tokens on more complex tasks (Task 2 and Task 3). Qwen3-thinking exhibits a different pattern, consistently using the most tokens across all tasks and achieving moderate performance in Task 2 and the best performance in Task 3.

\section{Case Study}

To provide a qualitative understanding of the models' reasoning processes, we analyze excerpts from the thinking traces of qwen3-thinking and gpt-oss-120b on one instance of Task 3. This analysis reveals stark differences in reasoning styles and efficiency.

\subsection{qwen3-thinking: A Verbose and Repetitive Process}

The reasoning trace for qwen3-thinking is characterized by its verbosity and a tendency to repeatedly re-state definitions and the primary goal. The trace begins with a lengthy preamble where the model outlines the task and defines the core concepts multiple times before processing the input.

\begin{tcolorbox}[
  enhanced,
  colback=purple!5,
  colframe=purple!75,
  boxrule=0.5pt,
  fonttitle=\bfseries,
  title=Excerpt 1: Repetitive Goal Setting,
]
First, I need to identify all significant distinctions between the current case (C1) and the precedent case (C2). [...] The key definitions:
\begin{itemize}
    \item {\color{orange60}A distinction is an unshared factor between C1 and C2. There are two types:
    \begin{enumerate}
        \item A factor favoring C2's winner (defendant) is present in C2 but absent in C1.
        \item A factor favoring C2's loser (plaintiff) is present in C1 but absent in C2.
    \end{enumerate}}
\end{itemize}
[...] So we have four distinctions: F11(d), F27(d), F15(p), F18(p). Now, for each of these, I need to check if it is significant. \textcolor{orange60}{A significant distinction: can be emphasized and cannot be downplayed. [...] Let me read carefully.} ``A distinction can be emphasized if...'' ``A distinction can be downplayed if...'' [...]
\end{tcolorbox}

When analyzing specific distinctions, the model proceeds with a step-by-step application of the rules. However, this process is again marked by verbosity. The excerpt below shows the model correctly concluding that distinction F27(d) is not significant because it can be downplayed, but the reasoning is exhaustive.

\begin{tcolorbox}[
  enhanced,
  colback=purple!5,
  colframe=purple!75,
  boxrule=0.5pt,
  fonttitle=\bfseries,
  title=Excerpt 2: Exhaustive Analysis of a Single Distinction,
]

[...] Now, for downplay: \textcolor{orange60}{a distinction can be downplayed if there is a concern/issue where F27 has support for defendant and C1 has alternative effective support for defendant.} For C102: F27 has support for defendant, and C1 also has support for defendant (from F1 via C122), so for C102, C1 has alternative support for defendant. [...] So for C102, downplay is possible. [...] Therefore, for F27(d), it can be emphasized [...] but it can be downplayed [...], so not significant. [...]
\end{tcolorbox}

This trace illustrates an inefficient reasoning style while ultimately capable of reaching the correct conclusions for intermediate steps. The model's tendency to ``think out loud'' about every detail, including redundant definitions, contributes to its high token consumption.

\subsection{gpt-oss-120b: A More Concise, but Flawed, Approach}

In contrast, the reasoning trace for gpt-oss-120b is more concise. The model quickly identifies the distinctions and proceeds to evaluate them without the extensive preamble observed in qwen3-thinking.

\begin{tcolorbox}[
  enhanced,
  colback=blue!5,
  colframe=blue!75,
  boxrule=0.5pt,
  fonttitle=\bfseries,
  title=Excerpt 1: Direct Identification of Distinctions,
]
[...] C2 winner is defendant. So "favoring C2's winner" = factors that favor defendant (d) present in C2 but not in C1. [...] distinctions: F11(d), F27(d) (type 1). Next plaintiff factors present in C1 but not in C2. [...] distinct plaintiff factors: F15(p), F18(p). Those are type 2 distinctions. Thus total distinctions: F11(d), F27(d), F15(p), F18(p). [...]
\end{tcolorbox}

However, the model's haste appears to cause a factual error. It fails to identify a key support path in the hierarchy for one of the concerns, leading it to an incorrect conclusion.

\begin{tcolorbox}[
  enhanced,
  colback=blue!5,
  colframe=blue!75,
  boxrule=0.5pt,
  fonttitle=\bfseries,
  title=Excerpt 2: Factual Error due to Missed Path,
]
[...] Distinction D: F18(p) – effective support for I112 and I110. \textcolor{cyan60}{Check C2 effective support for I112? C2 factors: none linking to I112. [...] So C2 lacks support for I112. Also for I110 (downstream via I112). C2 lacks support.} Any alternative support for I112 or I110 from C2? No. So no downplay. Thus F18(p) seems emphasized [...] and not downplayed. So it's significant. [...]
\end{tcolorbox}

The model missed that in C2, factor F22(p) provides strong support for I110 via concern C111. This oversight led it to incorrectly identify F18(p) as significant. This comparison reveals a trade-off: qwen3-thinking's verbose process, while inefficient, was more methodical and led to a more accurate application of the rules. In contrast, gpt-oss-120b's faster, more direct approach was brittle and resulted in reasoning failures. This suggests that for complex, rule-based reasoning, a more deliberative process may be necessary to ensure accuracy.

\section{Discussion}

\begin{figure}[t]
\centering
\includegraphics[width=0.95\columnwidth]{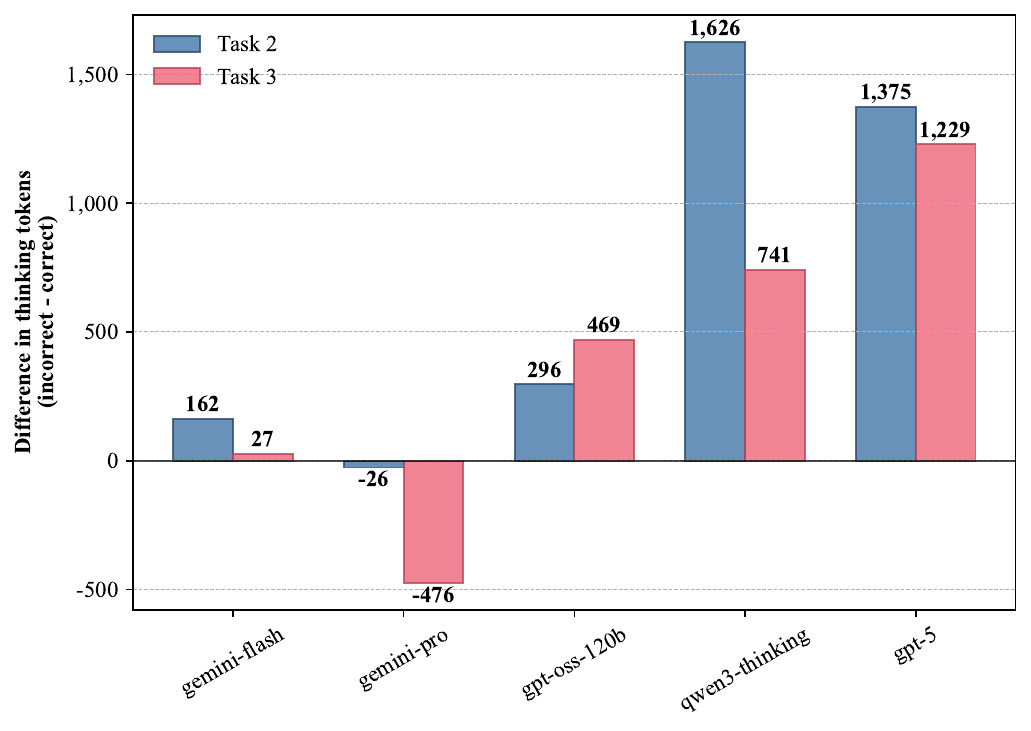}
\caption{Token usage patterns reveal inefficient reasoning strategies. The figure shows the difference in thinking tokens between incorrect and correct responses across Tasks 2--3, highlighting how models often expend more computational effort on answers they ultimately get wrong.}
\label{fig:thinking_differences}
\end{figure}

Our findings reveal insights on the relationship between computational effort and reasoning quality in current reasoning LLMs.

One of the findings is the consistent pattern where models expend more computational resources on incorrect responses than correct ones. As shown in Figure \ref{fig:thinking_differences} and detailed in Table \ref{tab:thinking-tokens-by-correctness}, this phenomenon is observed across multiple models and tasks. For example, gpt-5 used 4,456 tokens on average for incorrect Task 2 responses compared to 3,081 for correct ones, which was a 45\% increase in computational effort for worse outcomes.

This pattern suggests that when models encounter difficult problems, they may fall into inefficient reasoning loops or engage in extensive but unfocused analysis. Rather than converging on the correct solution, they appear to ``overthink'' the problem, exploring multiple reasoning paths without successfully identifying the most effective approach.

The consistency of this pattern across different models suggests that it may represent a limitation in current LLM reasoning mechanisms and the corresponding post-training approaches. The models appear to lack effective strategies for recognizing when they are on the wrong track and redirecting their reasoning efforts accordingly.

A second important finding is the lack of strong correlation between performance and reasoning effort across models. Qwen3-thinking exemplifies this disconnect between performance and reasoning effort, using 9,596 tokens on average for Task 2 while achieving only 78.66\% accuracy, compared to gpt-5's 3,189 tokens and 92.09\% accuracy. This represents a 3.0-fold increase in computational effort for worse performance.

Gpt-5 demonstrates the most efficient reasoning in this study, achieving high accuracy with moderate token usage. Gemini-pro shows an interesting pattern where it uses fewer tokens than gemini-flash on Task 1, it achieves higher accuracy with more tokens on Task 2 and Task 3.

This finding challenges the assumption that more reasoning efforts leads to better outcomes. Instead, it suggests that the quality of reasoning is more important than its quantity. Models that can identify and follow effective reasoning paths achieve better results with less computational overhead than those that engage in extensive but unfocused analysis. Simply increasing the reasoning length may not improve their performance on reasoning tasks.

\section{Conclusion}

This work presents a systematic investigation into the capabilities and limitations of LLMs in hierarchical legal reasoning, specifically focusing on the task of identifying significant distinctions between legal cases.

Our primary contribution is the development of a formal, three-stage framework that decomposes the process of identifying significant distinctions into increasingly complex reasoning tasks. The formal representation in our framework exposes a core limitation in LLMs' hierarchical reasoning. This same limitation may exist but be obfuscated when LLMs operate on contextual narrative legal work, especially in areas well-represented in their pre-training data or accessible via retrieval-augmented generation. In such cases, models might appear to reason effectively by retrieving and adapting existing solutions, masking the underlying deficit in hierarchical problem-solving.

This decomposition reveals that while LLMs excel at surface-level pattern recognition (Task 1: 100\% accuracy across all models), their performance degrades when required to engage in hierarchical reasoning about argumentative roles (Task 2: 64.82\%-92.09\%) and collapses when synthesizing multiple reasoning steps (Task 3: 11.46\%-33.99\%). This performance gradient provides insights into the specific failure modes of current LLMs in legal reasoning.

As demonstrated by the comparison between thinking and non-thinking models, the introduction of reasoning capabilities improves the accuracy. The qwen3-thinking model achieved 78.66\% accuracy on Task 2 compared to 30.04\% for its non-thinking counterpart, and maintained 33.99\% accuracy on Task 3 where the non-thinking model failed. This finding demonstrates that the reasoning capabilities enabled by the reinforcement-learning-based post-training improves the accuracy on hierarchical reasoning tasks.

Our analysis reveals a disconnect between the quantity and quality of reasoning in current LLMs. Models consistently expend more computational resources on incorrect responses than correct ones, with gpt-5 using 4,456 tokens on average for incorrect Task 2 responses compared to 3,081 for correct ones. This pattern suggests that models may fall into inefficient reasoning loops when encountering difficult problems. The relationship between reasoning effort and performance varies across models. Qwen3-thinking uses 3.0 times more tokens than gpt-5 while achieving lower accuracy on Task 2. These findings suggest that the next frontier in LLM development lies not only in longer reasoning chains, but also in elevating the caliber of training data and refining the models' reasoning abilities in various domains.

The systematic failure of LLMs on integrated reasoning tasks (Task 3) has implications for the deployment of AI systems in legal contexts. The ability to identify significant distinctions is important to legal practice, and the current inability of models to perform this task reliably raises concerns about their readiness for real-world legal applications. Our framework provides a methodology for identifying and addressing these limitations before deployment. The hierarchical nature of legal reasoning, as captured by our CATO-based framework, reveals that effective legal AI must be capable of multi-step, multi-level abstraction and synthesis. The current generation of LLMs, while impressive in their surface-level capabilities, lacks the sophisticated reasoning mechanisms necessary for the nuanced analysis required in legal practice.

\section{Limitations and Future Work}

Our evaluation is currently limited to the domain of trade secret law within the U.S. legal system, as represented by the CATO framework. This focus, while providing a controlled environment for systematic evaluation, may not capture the full complexity and diversity of legal reasoning across different jurisdictions and legal domains. The hierarchical structures, reasoning patterns, and argumentative strategies may vary across different areas of law.

The systematic failures we observe in integrated reasoning tasks (Task 3) suggest the need for innovative approaches to LLM post-training. Current reasoning LLMs are typically trained using reinforcement learning on mathematical and logical reasoning tasks, which may not adequately prepare them for the domain-specific reasoning patterns required in legal analysis. The domain shift from mathematical and logical reasoning to legal analysis may impact models' performance in this study. Developing specialized training datasets and post-training methodologies designed for legal reasoning, with careful attention to the hierarchical and multi-step nature of legal argumentation, might improve their reasoning capabilities in legal context.

Our current framework provides a solid foundation, but several enhancements could improve its utility and scope. Future work should develop adaptive difficulty mechanisms that adjust task complexity based on model performance, providing more nuanced insights into capability boundaries. Incorporating more human legal experts in the evaluation loop would provide insights into the practical significance of model performance in real-world legal contexts.

\bibliographystyle{ACM-Reference-Format}
\bibliography{reference}


\begin{thebibliography}{56}


\ifx \showCODEN    \undefined \def \showCODEN     #1{\unskip}     \fi
\ifx \showISBNx    \undefined \def \showISBNx     #1{\unskip}     \fi
\ifx \showISBNxiii \undefined \def \showISBNxiii  #1{\unskip}     \fi
\ifx \showISSN     \undefined \def \showISSN      #1{\unskip}     \fi
\ifx \showLCCN     \undefined \def \showLCCN      #1{\unskip}     \fi
\ifx \shownote     \undefined \def \shownote      #1{#1}          \fi
\ifx \showarticletitle \undefined \def \showarticletitle #1{#1}   \fi
\ifx \showURL      \undefined \def \showURL       {\relax}        \fi
\providecommand\bibfield[2]{#2}
\providecommand\bibinfo[2]{#2}
\providecommand\natexlab[1]{#1}
\providecommand\showeprint[2][]{arXiv:#2}

\bibitem[Afroogh et~al\mbox{.}(2024)]%
        {afroogh2024trust}
\bibfield{author}{\bibinfo{person}{Saleh Afroogh}, \bibinfo{person}{Ali Akbari}, \bibinfo{person}{Emmie Malone}, \bibinfo{person}{Mohammadali Kargar}, {and} \bibinfo{person}{Hananeh Alambeigi}.} \bibinfo{year}{2024}\natexlab{}.
\newblock \showarticletitle{Trust in AI: progress, challenges, and future directions}.
\newblock \bibinfo{journal}{\emph{Humanities and Social Sciences Communications}} \bibinfo{volume}{11}, \bibinfo{number}{1} (\bibinfo{year}{2024}), \bibinfo{pages}{1--30}.
\newblock


\bibitem[Agarwal et~al\mbox{.}(2025)]%
        {agarwal2025gpt}
\bibfield{author}{\bibinfo{person}{Sandhini Agarwal}, \bibinfo{person}{Lama Ahmad}, \bibinfo{person}{Jason Ai}, \bibinfo{person}{Sam Altman}, \bibinfo{person}{Andy Applebaum}, \bibinfo{person}{Edwin Arbus}, \bibinfo{person}{Rahul~K Arora}, \bibinfo{person}{Yu Bai}, \bibinfo{person}{Bowen Baker}, \bibinfo{person}{Haiming Bao}, {et~al\mbox{.}}} \bibinfo{year}{2025}\natexlab{}.
\newblock \showarticletitle{gpt-oss-120b \& gpt-oss-20b model card}.
\newblock \bibinfo{journal}{\emph{arXiv preprint arXiv:2508.10925}} (\bibinfo{year}{2025}).
\newblock


\bibitem[Aletras et~al\mbox{.}(2016)]%
        {aletras2016predicting}
\bibfield{author}{\bibinfo{person}{Nikolaos Aletras}, \bibinfo{person}{Dimitrios Tsarapatsanis}, \bibinfo{person}{Daniel Preo{\c{t}}iuc-Pietro}, {and} \bibinfo{person}{Vasileios Lampos}.} \bibinfo{year}{2016}\natexlab{}.
\newblock \showarticletitle{Predicting judicial decisions of the European Court of Human Rights: A natural language processing perspective}.
\newblock \bibinfo{journal}{\emph{PeerJ computer science}}  \bibinfo{volume}{2} (\bibinfo{year}{2016}), \bibinfo{pages}{e93}.
\newblock


\bibitem[Aleven and Ashley(1995)]%
        {aleven1995doing}
\bibfield{author}{\bibinfo{person}{Vincent Aleven} {and} \bibinfo{person}{Kevin~D Ashley}.} \bibinfo{year}{1995}\natexlab{}.
\newblock \showarticletitle{Doing things with factors}. In \bibinfo{booktitle}{\emph{Proceedings of the 5th international conference on artificial intelligence and law}}. \bibinfo{pages}{31--41}.
\newblock


\bibitem[Aleven(1997)]%
        {aleven1997teaching}
\bibfield{author}{\bibinfo{person}{Vincent~AWMM Aleven}.} \bibinfo{year}{1997}\natexlab{}.
\newblock \bibinfo{booktitle}{\emph{Teaching case-based argumentation through a model and examples}}.
\newblock \bibinfo{publisher}{University of Pittsburgh}.
\newblock


\bibitem[Ashley(1991)]%
        {ashley1991reasoning}
\bibfield{author}{\bibinfo{person}{Kevin~D Ashley}.} \bibinfo{year}{1991}\natexlab{}.
\newblock \showarticletitle{Reasoning with cases and hypotheticals in HYPO}.
\newblock \bibinfo{journal}{\emph{International journal of man-machine studies}} \bibinfo{volume}{34}, \bibinfo{number}{6} (\bibinfo{year}{1991}), \bibinfo{pages}{753--796}.
\newblock


\bibitem[Ashley(1992)]%
        {ashley1992case}
\bibfield{author}{\bibinfo{person}{Kevin~D Ashley}.} \bibinfo{year}{1992}\natexlab{}.
\newblock \showarticletitle{Case-based reasoning and its implications for legal expert systems}.
\newblock \bibinfo{journal}{\emph{Artificial Intelligence and Law}} \bibinfo{volume}{1}, \bibinfo{number}{2} (\bibinfo{year}{1992}), \bibinfo{pages}{113--208}.
\newblock


\bibitem[Ashley(2017)]%
        {ashley2017artificial}
\bibfield{author}{\bibinfo{person}{Kevin~D Ashley}.} \bibinfo{year}{2017}\natexlab{}.
\newblock \bibinfo{booktitle}{\emph{Artificial intelligence and legal analytics: new tools for law practice in the digital age}}.
\newblock \bibinfo{publisher}{Cambridge University Press}.
\newblock


\bibitem[Ashley and Aleven(1997)]%
        {ashley1997reasoning}
\bibfield{author}{\bibinfo{person}{Kevin~D Ashley} {and} \bibinfo{person}{Vincent Aleven}.} \bibinfo{year}{1997}\natexlab{}.
\newblock \showarticletitle{Reasoning symbolically about partially matched cases}. In \bibinfo{booktitle}{\emph{IJCAI (1)}}. \bibinfo{pages}{335--341}.
\newblock


\bibitem[Atkinson and Bench-Capon(2019)]%
        {atkinson2019reasoning}
\bibfield{author}{\bibinfo{person}{Katie Atkinson} {and} \bibinfo{person}{Trevor Bench-Capon}.} \bibinfo{year}{2019}\natexlab{}.
\newblock \showarticletitle{Reasoning with legal cases: Analogy or rule application?}. In \bibinfo{booktitle}{\emph{Proceedings of the Seventeenth International Conference on Artificial Intelligence and Law}}. \bibinfo{pages}{12--21}.
\newblock


\bibitem[Bilal et~al\mbox{.}(2025)]%
        {bilal2025llms}
\bibfield{author}{\bibinfo{person}{Ahsan Bilal}, \bibinfo{person}{David Ebert}, {and} \bibinfo{person}{Beiyu Lin}.} \bibinfo{year}{2025}\natexlab{}.
\newblock \showarticletitle{Llms for explainable ai: A comprehensive survey}.
\newblock \bibinfo{journal}{\emph{arXiv preprint arXiv:2504.00125}} (\bibinfo{year}{2025}).
\newblock


\bibitem[Br{\"u}ninghaus and Ashley(2003)]%
        {bruninghaus2003predicting}
\bibfield{author}{\bibinfo{person}{Stefanie Br{\"u}ninghaus} {and} \bibinfo{person}{Kevin~D Ashley}.} \bibinfo{year}{2003}\natexlab{}.
\newblock \showarticletitle{Predicting outcomes of case based legal arguments}. In \bibinfo{booktitle}{\emph{Proceedings of the 9th International conference on Artificial Intelligence and Law}}. ACM, \bibinfo{pages}{233--242}.
\newblock


\bibitem[Buhl et~al\mbox{.}(2025)]%
        {buhl2025alignment}
\bibfield{author}{\bibinfo{person}{Marie~Davidsen Buhl}, \bibinfo{person}{Jacob Pfau}, \bibinfo{person}{Benjamin Hilton}, {and} \bibinfo{person}{Geoffrey Irving}.} \bibinfo{year}{2025}\natexlab{}.
\newblock \showarticletitle{An alignment safety case sketch based on debate}.
\newblock \bibinfo{journal}{\emph{arXiv preprint arXiv:2505.03989}} (\bibinfo{year}{2025}).
\newblock


\bibitem[Canavotto and Horty(2023)]%
        {canavotto2023reasoning}
\bibfield{author}{\bibinfo{person}{Ilaria Canavotto} {and} \bibinfo{person}{John Horty}.} \bibinfo{year}{2023}\natexlab{}.
\newblock \showarticletitle{Reasoning with hierarchies of open-textured predicates}. In \bibinfo{booktitle}{\emph{Proceedings of the Nineteenth International Conference on Artificial Intelligence and Law}}. \bibinfo{pages}{52--61}.
\newblock


\bibitem[Chalkidis et~al\mbox{.}(2018)]%
        {chalkidis2018obligation}
\bibfield{author}{\bibinfo{person}{Ilias Chalkidis}, \bibinfo{person}{Ion Androutsopoulos}, {and} \bibinfo{person}{Achilleas Michos}.} \bibinfo{year}{2018}\natexlab{}.
\newblock \showarticletitle{Obligation and prohibition extraction using hierarchical RNNs}.
\newblock \bibinfo{journal}{\emph{arXiv preprint arXiv:1805.03871}} (\bibinfo{year}{2018}).
\newblock


\bibitem[Chau and Livermore(2024)]%
        {chau2024computational}
\bibfield{author}{\bibinfo{person}{Bao~Kham Chau} {and} \bibinfo{person}{Michael~A Livermore}.} \bibinfo{year}{2024}\natexlab{}.
\newblock \showarticletitle{Computational legal studies comes of age}.
\newblock  (\bibinfo{year}{2024}).
\newblock


\bibitem[Chhatwal et~al\mbox{.}(2018)]%
        {chhatwal2018explainable}
\bibfield{author}{\bibinfo{person}{Rishi Chhatwal}, \bibinfo{person}{Peter Gronvall}, \bibinfo{person}{Nathaniel Huber-Fliflet}, \bibinfo{person}{Robert Keeling}, \bibinfo{person}{Jianping Zhang}, {and} \bibinfo{person}{Haozhen Zhao}.} \bibinfo{year}{2018}\natexlab{}.
\newblock \showarticletitle{Explainable text classification in legal document review a case study of explainable predictive coding}. In \bibinfo{booktitle}{\emph{2018 IEEE international conference on big data (Big Data)}}. IEEE, \bibinfo{pages}{1905--1911}.
\newblock


\bibitem[Collenette et~al\mbox{.}(2020)]%
        {collenette2020explainable}
\bibfield{author}{\bibinfo{person}{Joe Collenette}, \bibinfo{person}{Katie Atkinson}, {and} \bibinfo{person}{Trevor Bench-Capon}.} \bibinfo{year}{2020}\natexlab{}.
\newblock \showarticletitle{An explainable approach to deducing outcomes in European Court of Human Rights cases using ADFs}.
\newblock In \bibinfo{booktitle}{\emph{Computational Models of Argument}}. \bibinfo{publisher}{IOS Press}, \bibinfo{pages}{21--32}.
\newblock


\bibitem[Collenette et~al\mbox{.}(2023)]%
        {collenette2023explainable}
\bibfield{author}{\bibinfo{person}{Joe Collenette}, \bibinfo{person}{Katie Atkinson}, {and} \bibinfo{person}{Trevor Bench-Capon}.} \bibinfo{year}{2023}\natexlab{}.
\newblock \showarticletitle{Explainable AI tools for legal reasoning about cases: A study on the European Court of Human Rights}.
\newblock \bibinfo{journal}{\emph{Artificial Intelligence}}  \bibinfo{volume}{317} (\bibinfo{year}{2023}), \bibinfo{pages}{103861}.
\newblock


\bibitem[Comanici et~al\mbox{.}(2025)]%
        {comanici2025gemini}
\bibfield{author}{\bibinfo{person}{Gheorghe Comanici}, \bibinfo{person}{Eric Bieber}, \bibinfo{person}{Mike Schaekermann}, \bibinfo{person}{Ice Pasupat}, \bibinfo{person}{Noveen Sachdeva}, \bibinfo{person}{Inderjit Dhillon}, \bibinfo{person}{Marcel Blistein}, \bibinfo{person}{Ori Ram}, \bibinfo{person}{Dan Zhang}, \bibinfo{person}{Evan Rosen}, {et~al\mbox{.}}} \bibinfo{year}{2025}\natexlab{}.
\newblock \showarticletitle{Gemini 2.5: Pushing the frontier with advanced reasoning, multimodality, long context, and next generation agentic capabilities}.
\newblock \bibinfo{journal}{\emph{arXiv preprint arXiv:2507.06261}} (\bibinfo{year}{2025}).
\newblock


\bibitem[Douglas(1949)]%
        {douglas1949stare}
\bibfield{author}{\bibinfo{person}{William~O Douglas}.} \bibinfo{year}{1949}\natexlab{}.
\newblock \showarticletitle{Stare decisis}.
\newblock \bibinfo{journal}{\emph{Columbia Law Review}} \bibinfo{volume}{49}, \bibinfo{number}{6} (\bibinfo{year}{1949}), \bibinfo{pages}{735--758}.
\newblock


\bibitem[Gerhardt(2011)]%
        {gerhardt2011power}
\bibfield{author}{\bibinfo{person}{Michael~J Gerhardt}.} \bibinfo{year}{2011}\natexlab{}.
\newblock \bibinfo{booktitle}{\emph{The power of precedent}}.
\newblock \bibinfo{publisher}{Oxford University Press}.
\newblock


\bibitem[Goebel et~al\mbox{.}(2025)]%
        {goebel2025international}
\bibfield{author}{\bibinfo{person}{Randy Goebel}, \bibinfo{person}{Yoshinobu Kano}, \bibinfo{person}{Japan~Calum Kawn}, \bibinfo{person}{Mi-Young Kim}, {and} \bibinfo{person}{Masaharu Yoshioka}.} \bibinfo{year}{2025}\natexlab{}.
\newblock \showarticletitle{International Competition on Legal Information Extraction and Entailment (COLIEE 2025)}.
\newblock  (\bibinfo{year}{2025}).
\newblock


\bibitem[Grabmair(2016)]%
        {grabmair2016modeling}
\bibfield{author}{\bibinfo{person}{Matthias Grabmair}.} \bibinfo{year}{2016}\natexlab{}.
\newblock \emph{\bibinfo{title}{Modeling purposive legal argumentation and case outcome prediction using argument schemes in the value judgment formalism}}.
\newblock \bibinfo{thesistype}{Ph.\,D. Dissertation}. \bibinfo{school}{University of Pittsburgh}.
\newblock


\bibitem[Grabmair(2017)]%
        {grabmair2017predicting}
\bibfield{author}{\bibinfo{person}{Matthias Grabmair}.} \bibinfo{year}{2017}\natexlab{}.
\newblock \showarticletitle{Predicting trade secret case outcomes using argument schemes and learned quantitative value effect tradeoffs}. In \bibinfo{booktitle}{\emph{Proceedings of the 16th edition of the International Conference on Articial Intelligence and Law}}. \bibinfo{pages}{89--98}.
\newblock


\bibitem[Gray et~al\mbox{.}(2025)]%
        {gray2025generating}
\bibfield{author}{\bibinfo{person}{Morgan Gray}, \bibinfo{person}{Li Zhang}, {and} \bibinfo{person}{Kevin~D Ashley}.} \bibinfo{year}{2025}\natexlab{}.
\newblock \showarticletitle{Generating case-based legal arguments with llms}. In \bibinfo{booktitle}{\emph{Proceedings of the 2025 Symposium on Computer Science and Law}}. \bibinfo{pages}{160--168}.
\newblock


\bibitem[Guha et~al\mbox{.}(2023)]%
        {guha2023legalbench}
\bibfield{author}{\bibinfo{person}{Neel Guha}, \bibinfo{person}{Julian Nyarko}, \bibinfo{person}{Daniel Ho}, \bibinfo{person}{Christopher R{\'e}}, \bibinfo{person}{Adam Chilton}, \bibinfo{person}{Alex Chohlas-Wood}, \bibinfo{person}{Austin Peters}, \bibinfo{person}{Brandon Waldon}, \bibinfo{person}{Daniel Rockmore}, \bibinfo{person}{Diego Zambrano}, {et~al\mbox{.}}} \bibinfo{year}{2023}\natexlab{}.
\newblock \showarticletitle{Legalbench: A collaboratively built benchmark for measuring legal reasoning in large language models}.
\newblock \bibinfo{journal}{\emph{Advances in neural information processing systems}}  \bibinfo{volume}{36} (\bibinfo{year}{2023}), \bibinfo{pages}{44123--44279}.
\newblock


\bibitem[Guo et~al\mbox{.}(2025)]%
        {guo2025deepseek}
\bibfield{author}{\bibinfo{person}{Daya Guo}, \bibinfo{person}{Dejian Yang}, \bibinfo{person}{Haowei Zhang}, \bibinfo{person}{Junxiao Song}, \bibinfo{person}{Ruoyu Zhang}, \bibinfo{person}{Runxin Xu}, \bibinfo{person}{Qihao Zhu}, \bibinfo{person}{Shirong Ma}, \bibinfo{person}{Peiyi Wang}, \bibinfo{person}{Xiao Bi}, {et~al\mbox{.}}} \bibinfo{year}{2025}\natexlab{}.
\newblock \showarticletitle{Deepseek-r1: Incentivizing reasoning capability in llms via reinforcement learning}.
\newblock \bibinfo{journal}{\emph{arXiv preprint arXiv:2501.12948}} (\bibinfo{year}{2025}).
\newblock


\bibitem[Hendrycks et~al\mbox{.}(2021)]%
        {hendrycks2021cuad}
\bibfield{author}{\bibinfo{person}{Dan Hendrycks}, \bibinfo{person}{Collin Burns}, \bibinfo{person}{Anya Chen}, {and} \bibinfo{person}{Spencer Ball}.} \bibinfo{year}{2021}\natexlab{}.
\newblock \showarticletitle{Cuad: An expert-annotated nlp dataset for legal contract review}.
\newblock \bibinfo{journal}{\emph{arXiv preprint arXiv:2103.06268}} (\bibinfo{year}{2021}).
\newblock


\bibitem[Horty(2011)]%
        {horty2011rules}
\bibfield{author}{\bibinfo{person}{John~F Horty}.} \bibinfo{year}{2011}\natexlab{}.
\newblock \showarticletitle{Rules and reasons in the theory of precedent}.
\newblock \bibinfo{journal}{\emph{Legal theory}} \bibinfo{volume}{17}, \bibinfo{number}{1} (\bibinfo{year}{2011}), \bibinfo{pages}{1--33}.
\newblock


\bibitem[Hubinger et~al\mbox{.}(2024)]%
        {hubinger2024sleeper}
\bibfield{author}{\bibinfo{person}{Evan Hubinger}, \bibinfo{person}{Carson Denison}, \bibinfo{person}{Jesse Mu}, \bibinfo{person}{Mike Lambert}, \bibinfo{person}{Meg Tong}, \bibinfo{person}{Monte MacDiarmid}, \bibinfo{person}{Tamera Lanham}, \bibinfo{person}{Daniel~M Ziegler}, \bibinfo{person}{Tim Maxwell}, \bibinfo{person}{Newton Cheng}, {et~al\mbox{.}}} \bibinfo{year}{2024}\natexlab{}.
\newblock \showarticletitle{Sleeper agents: Training deceptive llms that persist through safety training}.
\newblock \bibinfo{journal}{\emph{arXiv preprint arXiv:2401.05566}} (\bibinfo{year}{2024}).
\newblock


\bibitem[Irving et~al\mbox{.}(2018)]%
        {irving2018ai}
\bibfield{author}{\bibinfo{person}{Geoffrey Irving}, \bibinfo{person}{Paul Christiano}, {and} \bibinfo{person}{Dario Amodei}.} \bibinfo{year}{2018}\natexlab{}.
\newblock \showarticletitle{AI safety via debate}.
\newblock \bibinfo{journal}{\emph{arXiv preprint arXiv:1805.00899}} (\bibinfo{year}{2018}).
\newblock


\bibitem[Kattnig et~al\mbox{.}(2024)]%
        {kattnig2024assessing}
\bibfield{author}{\bibinfo{person}{Markus Kattnig}, \bibinfo{person}{Alessa Angerschmid}, \bibinfo{person}{Thomas Reichel}, {and} \bibinfo{person}{Roman Kern}.} \bibinfo{year}{2024}\natexlab{}.
\newblock \showarticletitle{Assessing trustworthy AI: Technical and legal perspectives of fairness in AI}.
\newblock \bibinfo{journal}{\emph{Computer Law \& Security Review}}  \bibinfo{volume}{55} (\bibinfo{year}{2024}), \bibinfo{pages}{106053}.
\newblock


\bibitem[Ke et~al\mbox{.}(2025)]%
        {ke2025survey}
\bibfield{author}{\bibinfo{person}{Zixuan Ke}, \bibinfo{person}{Fangkai Jiao}, \bibinfo{person}{Yifei Ming}, \bibinfo{person}{Xuan-Phi Nguyen}, \bibinfo{person}{Austin Xu}, \bibinfo{person}{Do~Xuan Long}, \bibinfo{person}{Minzhi Li}, \bibinfo{person}{Chengwei Qin}, \bibinfo{person}{Peifeng Wang}, \bibinfo{person}{Silvio Savarese}, {et~al\mbox{.}}} \bibinfo{year}{2025}\natexlab{}.
\newblock \showarticletitle{A survey of frontiers in llm reasoning: Inference scaling, learning to reason, and agentic systems}.
\newblock \bibinfo{journal}{\emph{arXiv preprint arXiv:2504.09037}} (\bibinfo{year}{2025}).
\newblock


\bibitem[Kim and Goebel(2017)]%
        {kim2017two}
\bibfield{author}{\bibinfo{person}{Mi-Young Kim} {and} \bibinfo{person}{Randy Goebel}.} \bibinfo{year}{2017}\natexlab{}.
\newblock \showarticletitle{Two-step cascaded textual entailment for legal bar exam question answering}. In \bibinfo{booktitle}{\emph{Proceedings of the 16th edition of the International Conference on Articial Intelligence and Law}}. \bibinfo{pages}{283--290}.
\newblock


\bibitem[Lamond(2006)]%
        {lamond2006precedent}
\bibfield{author}{\bibinfo{person}{Grant Lamond}.} \bibinfo{year}{2006}\natexlab{}.
\newblock \showarticletitle{Precedent and analogy in legal reasoning}.
\newblock  (\bibinfo{year}{2006}).
\newblock


\bibitem[Lyu et~al\mbox{.}(2023)]%
        {lyu2023multi}
\bibfield{author}{\bibinfo{person}{Yougang Lyu}, \bibinfo{person}{Jitai Hao}, \bibinfo{person}{Zihan Wang}, \bibinfo{person}{Kai Zhao}, \bibinfo{person}{Shen Gao}, \bibinfo{person}{Pengjie Ren}, \bibinfo{person}{Zhumin Chen}, \bibinfo{person}{Fang Wang}, {and} \bibinfo{person}{Zhaochun Ren}.} \bibinfo{year}{2023}\natexlab{}.
\newblock \showarticletitle{Multi-defendant legal judgment prediction via hierarchical reasoning}.
\newblock \bibinfo{journal}{\emph{arXiv preprint arXiv:2312.05762}} (\bibinfo{year}{2023}).
\newblock


\bibitem[Mialon et~al\mbox{.}(2023)]%
        {mialon2023augmented}
\bibfield{author}{\bibinfo{person}{Gr{\'e}goire Mialon}, \bibinfo{person}{Roberto Dess{\`\i}}, \bibinfo{person}{Maria Lomeli}, \bibinfo{person}{Christoforos Nalmpantis}, \bibinfo{person}{Ram Pasunuru}, \bibinfo{person}{Roberta Raileanu}, \bibinfo{person}{Baptiste Rozi{\`e}re}, \bibinfo{person}{Timo Schick}, \bibinfo{person}{Jane Dwivedi-Yu}, \bibinfo{person}{Asli Celikyilmaz}, {et~al\mbox{.}}} \bibinfo{year}{2023}\natexlab{}.
\newblock \showarticletitle{Augmented language models: a survey}.
\newblock \bibinfo{journal}{\emph{arXiv preprint arXiv:2302.07842}} (\bibinfo{year}{2023}).
\newblock


\bibitem[{OpenAI}(2025)]%
        {openai_gpt5_system_card}
\bibfield{author}{\bibinfo{person}{{OpenAI}}.} \bibinfo{year}{2025}\natexlab{}.
\newblock \bibinfo{title}{GPT-5 System Card}.
\newblock \bibinfo{howpublished}{\url{https://cdn.openai.com/gpt-5-system-card.pdf}}.
\newblock
\newblock
\shownote{Accessed on 2025-08-26}.


\bibitem[Posner and Saran(2025)]%
        {posner2025judge}
\bibfield{author}{\bibinfo{person}{Eric~A Posner} {and} \bibinfo{person}{Shivam Saran}.} \bibinfo{year}{2025}\natexlab{}.
\newblock \showarticletitle{Judge AI: Assessing large language models in judicial decision-making}.
\newblock \bibinfo{journal}{\emph{University of Chicago Coase-Sandor Institute for Law \& Economics Research Paper}} \bibinfo{number}{2503} (\bibinfo{year}{2025}).
\newblock


\bibitem[{Qwen Team}(2025a)]%
        {qwen3-235b-non-thinking}
\bibfield{author}{\bibinfo{person}{{Qwen Team}}.} \bibinfo{year}{2025}\natexlab{a}.
\newblock \bibinfo{title}{Qwen3-235B-A22B-Instruct-2507-FP8}.
\newblock
\urldef\tempurl%
\url{https://huggingface.co/unsloth/Qwen3-235B-A22B-Instruct-2507-GGUF}
\showURL{%
\tempurl}


\bibitem[{Qwen Team}(2025b)]%
        {qwen3-235b-thinking}
\bibfield{author}{\bibinfo{person}{{Qwen Team}}.} \bibinfo{year}{2025}\natexlab{b}.
\newblock \bibinfo{title}{Qwen3-235B-A22B-Thinking-2507-FP8}.
\newblock
\urldef\tempurl%
\url{https://huggingface.co/Qwen/Qwen3-235B-A22B-Thinking-2507-FP8}
\showURL{%
\tempurl}


\bibitem[Rissland and Skalak(1991)]%
        {rissland1991cabaret}
\bibfield{author}{\bibinfo{person}{Edwina~L Rissland} {and} \bibinfo{person}{David~B Skalak}.} \bibinfo{year}{1991}\natexlab{}.
\newblock \showarticletitle{CABARET: rule interpretation in a hybrid architecture}.
\newblock \bibinfo{journal}{\emph{International journal of man-machine studies}} \bibinfo{volume}{34}, \bibinfo{number}{6} (\bibinfo{year}{1991}), \bibinfo{pages}{839--887}.
\newblock


\bibitem[Schauer(2008)]%
        {schauer2008precedent}
\bibfield{author}{\bibinfo{person}{Frederick Schauer}.} \bibinfo{year}{2008}\natexlab{}.
\newblock \showarticletitle{Why precedent in law (and elsewhere) is not totally (or even substantially) about analogy}.
\newblock \bibinfo{journal}{\emph{Perspectives on Psychological Science}} \bibinfo{volume}{3}, \bibinfo{number}{6} (\bibinfo{year}{2008}), \bibinfo{pages}{454--460}.
\newblock


\bibitem[Schauer and Spellman(2017)]%
        {schauer2017analogy}
\bibfield{author}{\bibinfo{person}{Frederick Schauer} {and} \bibinfo{person}{Barbara~A Spellman}.} \bibinfo{year}{2017}\natexlab{}.
\newblock \showarticletitle{Analogy, expertise, and experience}.
\newblock \bibinfo{journal}{\emph{The University of Chicago Law Review}} (\bibinfo{year}{2017}), \bibinfo{pages}{249--268}.
\newblock


\bibitem[Sesodia et~al\mbox{.}(2025)]%
        {sesodia2025annocaselaw}
\bibfield{author}{\bibinfo{person}{Magnus Sesodia}, \bibinfo{person}{Alina Petrova}, \bibinfo{person}{John Armour}, \bibinfo{person}{Thomas Lukasiewicz}, \bibinfo{person}{Oana-Maria Camburu}, \bibinfo{person}{Puneet~K Dokania}, \bibinfo{person}{Philip Torr}, {and} \bibinfo{person}{Christian~Schroeder de Witt}.} \bibinfo{year}{2025}\natexlab{}.
\newblock \showarticletitle{AnnoCaseLaw: A Richly-Annotated Dataset For Benchmarking Explainable Legal Judgment Prediction}.
\newblock \bibinfo{journal}{\emph{arXiv preprint arXiv:2503.00128}} (\bibinfo{year}{2025}).
\newblock


\bibitem[Smith(2025)]%
        {smith2025history}
\bibfield{author}{\bibinfo{person}{Michael~L Smith}.} \bibinfo{year}{2025}\natexlab{}.
\newblock \showarticletitle{History as Precedent: Common Law Reasoning in Historical Investigation}.
\newblock \bibinfo{journal}{\emph{University of Pennsylvania Journal of Constitutional Law}} \bibinfo{volume}{27}, \bibinfo{number}{3} (\bibinfo{year}{2025}), \bibinfo{pages}{587}.
\newblock


\bibitem[{Unsloth AI}(2025)]%
        {unsloth_docs}
\bibfield{author}{\bibinfo{person}{{Unsloth AI}}.} \bibinfo{year}{2025}\natexlab{}.
\newblock \bibinfo{title}{gpt-oss: How to Run \& Fine-tune}.
\newblock
\urldef\tempurl%
\url{https://docs.unsloth.ai/new/gpt-oss-how-to-run-and-fine-tune#running-gpt-oss}
\showURL{%
\tempurl}


\bibitem[Wang et~al\mbox{.}(2025a)]%
        {wang2025hierarchical}
\bibfield{author}{\bibinfo{person}{Guan Wang}, \bibinfo{person}{Jin Li}, \bibinfo{person}{Yuhao Sun}, \bibinfo{person}{Xing Chen}, \bibinfo{person}{Changling Liu}, \bibinfo{person}{Yue Wu}, \bibinfo{person}{Meng Lu}, \bibinfo{person}{Sen Song}, {and} \bibinfo{person}{Yasin~Abbasi Yadkori}.} \bibinfo{year}{2025}\natexlab{a}.
\newblock \showarticletitle{Hierarchical Reasoning Model}.
\newblock \bibinfo{journal}{\emph{arXiv preprint arXiv:2506.21734}} (\bibinfo{year}{2025}).
\newblock


\bibitem[Wang et~al\mbox{.}(2025b)]%
        {wang2025survey}
\bibfield{author}{\bibinfo{person}{Peng-Yuan Wang}, \bibinfo{person}{Tian-Shuo Liu}, \bibinfo{person}{Chenyang Wang}, \bibinfo{person}{Yi-Di Wang}, \bibinfo{person}{Shu Yan}, \bibinfo{person}{Cheng-Xing Jia}, \bibinfo{person}{Xu-Hui Liu}, \bibinfo{person}{Xin-Wei Chen}, \bibinfo{person}{Jia-Cheng Xu}, \bibinfo{person}{Ziniu Li}, {et~al\mbox{.}}} \bibinfo{year}{2025}\natexlab{b}.
\newblock \showarticletitle{A Survey on Large Language Models for Mathematical Reasoning}.
\newblock \bibinfo{journal}{\emph{arXiv preprint arXiv:2506.08446}} (\bibinfo{year}{2025}).
\newblock


\bibitem[Wei et~al\mbox{.}(2025)]%
        {wei2025plangenllms}
\bibfield{author}{\bibinfo{person}{Hui Wei}, \bibinfo{person}{Zihao Zhang}, \bibinfo{person}{Shenghua He}, \bibinfo{person}{Tian Xia}, \bibinfo{person}{Shijia Pan}, {and} \bibinfo{person}{Fei Liu}.} \bibinfo{year}{2025}\natexlab{}.
\newblock \showarticletitle{Plangenllms: A modern survey of llm planning capabilities}.
\newblock \bibinfo{journal}{\emph{arXiv preprint arXiv:2502.11221}} (\bibinfo{year}{2025}).
\newblock


\bibitem[Wei et~al\mbox{.}(2022)]%
        {wei2022chain}
\bibfield{author}{\bibinfo{person}{Jason Wei}, \bibinfo{person}{Xuezhi Wang}, \bibinfo{person}{Dale Schuurmans}, \bibinfo{person}{Maarten Bosma}, \bibinfo{person}{Fei Xia}, \bibinfo{person}{Ed Chi}, \bibinfo{person}{Quoc~V Le}, \bibinfo{person}{Denny Zhou}, {et~al\mbox{.}}} \bibinfo{year}{2022}\natexlab{}.
\newblock \showarticletitle{Chain-of-thought prompting elicits reasoning in large language models}.
\newblock \bibinfo{journal}{\emph{Advances in neural information processing systems}}  \bibinfo{volume}{35} (\bibinfo{year}{2022}), \bibinfo{pages}{24824--24837}.
\newblock


\bibitem[Yang et~al\mbox{.}(2025)]%
        {yang2025qwen3}
\bibfield{author}{\bibinfo{person}{An Yang}, \bibinfo{person}{Anfeng Li}, \bibinfo{person}{Baosong Yang}, \bibinfo{person}{Beichen Zhang}, \bibinfo{person}{Binyuan Hui}, \bibinfo{person}{Bo Zheng}, \bibinfo{person}{Bowen Yu}, \bibinfo{person}{Chang Gao}, \bibinfo{person}{Chengen Huang}, \bibinfo{person}{Chenxu Lv}, {et~al\mbox{.}}} \bibinfo{year}{2025}\natexlab{}.
\newblock \showarticletitle{Qwen3 technical report}.
\newblock \bibinfo{journal}{\emph{arXiv preprint arXiv:2505.09388}} (\bibinfo{year}{2025}).
\newblock


\bibitem[Yao et~al\mbox{.}(2023)]%
        {yao2023tree}
\bibfield{author}{\bibinfo{person}{Shunyu Yao}, \bibinfo{person}{Dian Yu}, \bibinfo{person}{Jeffrey Zhao}, \bibinfo{person}{Izhak Shafran}, \bibinfo{person}{Tom Griffiths}, \bibinfo{person}{Yuan Cao}, {and} \bibinfo{person}{Karthik Narasimhan}.} \bibinfo{year}{2023}\natexlab{}.
\newblock \showarticletitle{Tree of thoughts: Deliberate problem solving with large language models}.
\newblock \bibinfo{journal}{\emph{Advances in neural information processing systems}}  \bibinfo{volume}{36} (\bibinfo{year}{2023}), \bibinfo{pages}{11809--11822}.
\newblock


\bibitem[Zhang and Ashley(2025)]%
        {zhang2025mitigating}
\bibfield{author}{\bibinfo{person}{Li Zhang} {and} \bibinfo{person}{Kevin~D Ashley}.} \bibinfo{year}{2025}\natexlab{}.
\newblock \showarticletitle{Mitigating Manipulation and Enhancing Persuasion: A Reflective Multi-Agent Approach for Legal Argument Generation}.
\newblock \bibinfo{journal}{\emph{arXiv preprint arXiv:2506.02992}} (\bibinfo{year}{2025}).
\newblock


\bibitem[Zhang et~al\mbox{.}(2025)]%
        {zhang2025measuring}
\bibfield{author}{\bibinfo{person}{Li Zhang}, \bibinfo{person}{Morgan Gray}, \bibinfo{person}{Jaromir Savelka}, {and} \bibinfo{person}{Kevin~D Ashley}.} \bibinfo{year}{2025}\natexlab{}.
\newblock \showarticletitle{Measuring Faithfulness and Abstention: An Automated Pipeline for Evaluating LLM-Generated 3-ply Case-Based Legal Arguments}.
\newblock \bibinfo{journal}{\emph{arXiv preprint arXiv:2506.00694}} (\bibinfo{year}{2025}).
\newblock


\end{thebibliography}

\section*{Appendix}

\appendix

\section{More Experimental Results}

Table \ref{tab:accuracy-thinking-tokens} and Table \ref{tab:thinking-tokens-by-correctness} provide the detailed results of model performance and token usage across the tasks.

\begin{table*}[h]
\centering
\small
\caption{Accuracy and thinking tokens across Tasks 1--3. Accuracy (higher is better) and average thinking tokens per instance.}
\begin{tabular}{l|ccc|ccc}
\toprule
\textbf{Model} & \multicolumn{3}{c|}{\textbf{Accuracy (\%)}} & \multicolumn{3}{c}{\textbf{Thinking Tokens}} \\
& \textbf{Task 1} & \textbf{Task 2} & \textbf{Task 3} & \textbf{Task 1} & \textbf{Task 2} & \textbf{Task 3} \\
\midrule
\textbf{gemini-flash} & 100.00 & 64.82 & 11.46 & 654.90 & 5,144.72 & 15,262.17 \\
\textbf{gemini-pro} & 100.00 & 85.77 & 21.74 & 929.30 & 2,973.59 & 7,019.41 \\
\textbf{gpt-oss-120b} & 100.00 & 77.08 & 22.53 & 356.56 & 1,642.34 & 3,942.67 \\
\textbf{qwen3-thinking} & 100.00 & 78.66 & 33.99 & 2010.80 & 9,596.17 & 15,677.82 \\
\textbf{gpt-5} & 100.00 & 92.09 & 23.32 & 487.01 & 3,189.27 & 7,025.01 \\
\textbf{qwen3-non-thinking} & 100.00 & 30.04 & 0.00 & / & / & / \\
\bottomrule
\end{tabular}
\label{tab:accuracy-thinking-tokens}
\end{table*}

\begin{table*}[h]
\centering
\small
\caption{Thinking tokens by correctness for Tasks 2 and 3.}
\begin{tabular}{l|cc|cc}
\toprule
\textbf{Model} & \multicolumn{2}{c|}{\textbf{Task 2}} & \multicolumn{2}{c}{\textbf{Task 3}} \\
& \textbf{Correct} & \textbf{Incorrect} & \textbf{Correct} & \textbf{Incorrect} \\
\midrule
\textbf{gemini-flash} & 5,087.61 & 5,249.65 & 15,263.26 & 15,290.12 \\
\textbf{gemini-pro} & 2,977.37 & 2,951.14 & 7,371.33 & 6,895.77 \\
\textbf{gpt-oss-120b} & 1,568.21 & 1,864.57 & 3,607.08 & 4,076.53 \\
\textbf{qwen3-thinking} & 9,246.05 & 10,872.13 & 15,392.91 & 16,133.68 \\
\textbf{gpt-5} & 3,081.44 & 4,456.33 & 6,126.01 & 7,354.63 \\
\bottomrule
\end{tabular}
\label{tab:thinking-tokens-by-correctness}
\end{table*}

\section{An Illustrative Example of the Decomposed Framework}
\label{sec:illustrative_example}

To illustrate our three-task framework, we use an example from U.S. trade secret law to walk through each task. For example, we can consider the following cases:

\begin{quote}
    \small
    \begin{itemize}
        \item \textbf{Current Case (C1)}: Company A (plaintiff) is suing a former employee (defendant) for trade secret theft. The employee knew the information was confidential. Company A lacked security measures, and they waived their employees' confidentiality agreement.
        \item \textbf{Precedent Case (C2)}: Company B (plaintiff) lost a trade secret lawsuit against a former employee (defendant) because they lacked security measures.
    \end{itemize}
\end{quote}

\paragraph{Factor Representation}
The natural language narratives can be translated into a set of pre-defined factors. A (d) marks a factor favoring the defendant, while (p) favors the plaintiff. For instance, ``lacked security measures'' corresponds to F19\_No-Security-Measures(d), which hurts the plaintiff's claim. Conversely, the employee knowing the information was confidential (F21\_Knew-Info-Confidential(p)) supports the plaintiff's argument, but waiving confidentiality agreement (F23\_Waiver-of-Confidentiality(d)) weakens their claim. The factor representations for our example cases are as follows:
\begin{itemize}
    \item \textbf{Current Case (C1) Factors}: F19\_No-Security-Measures(d), F21\_Knew-Info-Confidential(p), \newline F23\_Waiver-of-Confidentiality(d)
    \item \textbf{Precedent Case (C2) Factors}: F19\_No-Security-Measures(d)
    \item \textbf{Precedent (C2) Winner}: Defendant
\end{itemize}

\subsection{Task 1: Identify Distinctions}
With the cases represented as factors, the first task is to identify all unshared factors that could serve as distinctions.

\paragraph{Analysis of Distinctions}
A distinction is an unshared factor that indicates the precedent case (C2) is not a good analogy for the current case (C1). There are two types of distinctions:
\begin{enumerate}
    \item A factor favoring C2's winner is present in C2 but absent in C1.
    \item A factor favoring C2's loser is present in C1 but absent in C2.
\end{enumerate}

For our example:
\begin{itemize}
    \item \textbf{Question 1.1}: Are there any pro-defendant factors from C2 (the winner) that are absent in C1? C2 has F19\_No-Security-Measures(d) which favors the defendant. C1 also has F19\_No-Security-Measures(d). Therefore, there are no distinctions from this condition.
    \item \textbf{Question 1.2}: Are there any pro-plaintiff factors in C1 (favoring C2's loser) that were not in C2? Yes. C1 has F21\_Knew-Info-Confidential(p) which favors the plaintiff. C2 does not have this factor. Therefore, F21(p) is a distinction.
\end{itemize}
\textbf{Conclusion for Task 1}: The identified distinction is F21(p).

\subsection{Tasks 2 \& 3: Analyze and Identify Significant Distinctions}
The next steps involve using a legal knowledge hierarchy to determine which distinctions are significant.

\begin{figure*}[h]
    \centering
    \includegraphics[width=0.8\textwidth]{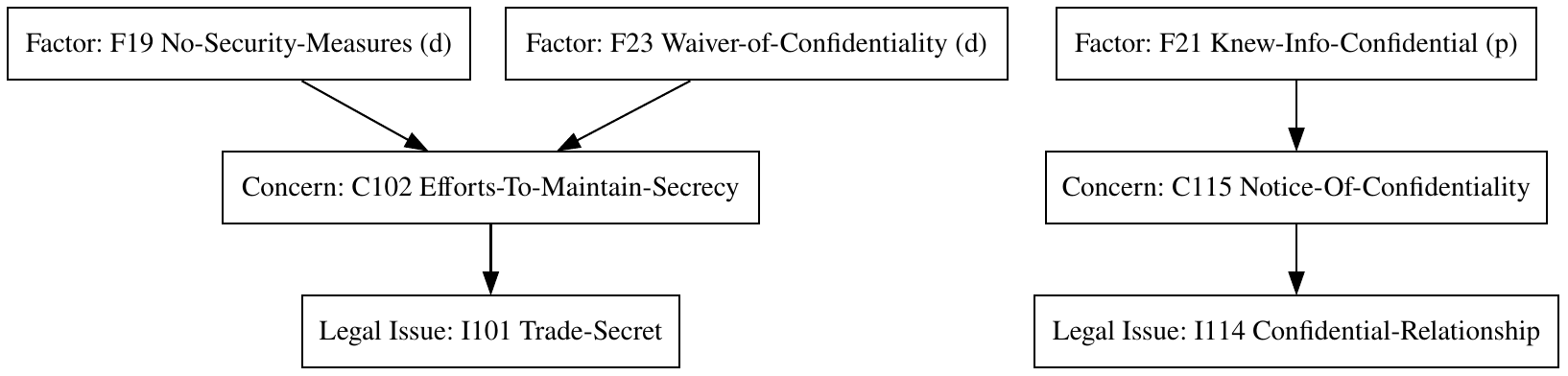}
    \caption{An example factor hierarchy for trade secret law, connecting base-level factors to intermediate legal concerns and top-level legal issues. Solid lines indicate strong support and dashed lines indicate weak support; this support can be for or against a concern, depending on which party the factor favors.}
    \label{fig:example_hierarchy}
\end{figure*}

\paragraph{The Factor Hierarchy} Factors provide evidence for more abstract legal concerns, which in turn inform high-level legal issues. In our example, we have two legal issues:
\begin{itemize}
    \item \textbf{Legal Issue I101}: Did the plaintiff's information constitute a trade secret?
    \begin{itemize}
        \item \textbf{Concern C102}: Did the plaintiff take efforts to maintain secrecy? This concern is supported or weakened by F23\_Waiver-of-Confidentiality(d) and F19\_No-Security-Measures(d).
    \end{itemize}
    \item \textbf{Legal Issue I114}: Was there a confidential relationship?
    \begin{itemize}
        \item \textbf{Concern C115}: Did the defendant have notice of confidentiality? This concern is supported by F21\_Knew-Info-Confidential(p).
    \end{itemize}
\end{itemize}
Figure \ref{fig:example_hierarchy} illustrates this structure.

\paragraph{Emphasis, Downplay, and Significance} The significance of a distinction depends on whether it can be emphasized or downplayed. A distinction is significant if it can be emphasized and cannot be downplayed.

\textbf{Definition of Downplaying}: A distinction $D$ can be downplayed if you can find an ancestor concern $P$ that meets both conditions: (1) $P$ is an ancestor of $D$ in the hierarchy, and (2) In C1, a different factor ($F_{\text{alt}}$) provides support for that same concern $P$.

\textbf{Definition of Emphasizing}: A distinction $D$ can be emphasized if you can find an ancestor concern $P$ that meets both conditions: (1) $P$ is an ancestor of $D$, and (2) The precedent case C2 lacks any factor that provides support for $P$ (when the distinction favors C2's loser) OR the current case C1 lacks any factor that provides support for $P$ (when the distinction favors C2's winner).

\begin{itemize}
    \item \textbf{Analysis of F21(p)}: F21(p) supports C115 \newline (Notice-Of-Confidentiality), which supports I114 (Confidential-Relationship). Since F21(p) favors C2's loser (plaintiff), we check if C2 lacks any factor providing support for C115. The Precedent Case (C2) has no pro-plaintiff factor supporting this concern. The absence of F21(p) in C2 is critical because it leaves this point entirely unsupported for the plaintiff in C2. Therefore, F21(p) \textbf{can be emphasized}. For downplaying, C1 has F21(p) as the only pro-plaintiff factor providing support for C115. While C1 also has F23\_Waiver-of-Confidentiality(d) and F19\_No-Security-Measures(d), these factors favor the defendant, not the plaintiff. Therefore, there is no alternative pro-plaintiff factor that can downplay the distinction. Thus, F21(p) \textbf{cannot be downplayed}. Table \ref{tab:significance_analysis} summarizes the significance analysis for this distinction.
\end{itemize}

\begin{tcolorbox}[enhanced, float=h, colback=blue!5, colframe=blue!75, boxrule=0.5pt, fonttitle=\bfseries, title=Legal Argument: Emphasizing a Distinction]
  \small
  \textbf{Plaintiff's Counsel in C1 argues:} ``Your Honor, the precedent case C2 is not a good analogy for our current case. In C2, the plaintiff had no factors to support the concern of notice of confidentiality (C115), which supports the legal issue of confidential relationship (I114). However, in our current case C1, we have F21\_Knew-Info-Confidential(p), which provides crucial support for C115. The presence of F21(p) in C1 strengthens our position regarding notice of confidentiality, making our case significantly stronger than the plaintiff's position was in C2. This distinction is critical and cannot be dismissed.''
  \end{tcolorbox}
  
  \begin{tcolorbox}[enhanced, float=h, colback=purple!5, colframe=purple!75, boxrule=0.5pt, fonttitle=\bfseries, title=Legal Argument: Failed Attempt to Downplay]
  \small
  \textbf{Defendant's Counsel in C1 would need to counter-argue:} To downplay this distinction, the defendant would need to show that the underlying concern C115 (Notice-Of-Confidentiality) is still addressed by alternative factors in C1. However, F21(p) is the only pro-plaintiff factor providing support for C115 in C1. While C1 has F23\_Waiver-of-Confidentiality(d) and F19\_No-Security-Measures(d), these factors favor the defendant, not the plaintiff. Therefore, there is no alternative pro-plaintiff factor that can downplay the distinction, leaving it successfully emphasized by the plaintiff.
  \end{tcolorbox}

\begin{table}[h]
\centering
\small
\caption{Significance Analysis of Distinctions}
\label{tab:significance_analysis}
\begin{tabularx}{\columnwidth}{@{}l>{\centering\arraybackslash}X>{\centering\arraybackslash}X>{\centering\arraybackslash}X@{}}
\toprule
\textbf{Distinction} & \textbf{Can be Emphasized?} & \textbf{Can be Downplayed?} & \textbf{Significant?} \\
\midrule
F21(p)   & YES & NO & \textbf{YES} \\
\bottomrule
\end{tabularx}
\end{table}

\textbf{Conclusion for Task 3}: The significant distinction is F21(p). The existence of this significant distinction suggests the precedent C2 should not apply in the current case C1.

\onecolumn

\section{Prompt Templates}
\label{sec:prompts}

This section provides the prompt templates used for each of the three tasks in our evaluation, as shown in Box \ref{box:task1}, Box \ref{box:task2}, and Box \ref{box:task3}.

\subsection{Task 1: Identify Distinctions}
\begin{tcolorbox}[enhanced, float=h, width=\columnwidth, title=Prompt for Task 1: Identify Distinctions, label=box:task1]
\small
\paragraph{Goal} Your task is to identify ALL distinctions between a current case (C1) and a precedent case (C2). This is the foundational step in legal case analysis.

\paragraph{Definition of Distinctions} A distinction is an unshared factor that indicates the precedent case (C2) is not a perfect analogy for the current case (C1). There are exactly two types of distinctions:
\begin{enumerate}
    \item \textbf{Type 1 Distinction}: A factor favoring C2's winner is present in C2 but absent in C1.
    \item \textbf{Type 2 Distinction}: A factor favoring C2's loser is present in C1 but absent in C2.
\end{enumerate}

\paragraph{Step-by-Step Process}
\begin{enumerate}
    \item Identify C2's winner and determine the winner and loser sides.
    \item Find all Type 1 distinctions by looking for factors present in C2, favoring the winner, and absent in C1.
    \item Find all Type 2 distinctions by looking for factors present in C1, favoring the loser, and absent in C2.
    \item Combine all distinctions found.
\end{enumerate}

\paragraph{One-shot Example} [...example...]

\paragraph{Output Format} Provide your analysis in a JSON object: `{"distinctions": ["F6(p)", "F19(d)", "F27(d)"]}`.
\end{tcolorbox}

\subsection{Task 2: Analyze Argumentative Roles of a Distinction}
\begin{tcolorbox}[enhanced, float=h, width=\columnwidth, title=Prompt for Task 2: Analyze a Single Distinction, label=box:task2]
\small
\paragraph{Goal} Your task is to identify whether a single given distinction is significant. Cases in this task are represented by factors. You will be given a hierarchy on the factors, concerns and issues in the Mermaid diagram, which is a structure where each node is a factor, concern or issue and edge(s) are directed paths between two nodes.

\paragraph{Key Definitions}
\begin{itemize}
    \item \textbf{Distinction}: [...definition...]
    \item \textbf{Factor Representation for Cases}: A factor is a pro-plaintiff (p) or pro-defendant (d) element that appears in the base level of the hierarchy. It is represented as F<number>(p) or F<number>(d). The middle level is the concern level (C) and the top level is the issue level (I).
    \item \textbf{Effective Support}: A factor's support for a concern/issue is effective if its path is strong (`-->`), or if its path is weak (`-.->`) and not blocked by an opposing factor's strong path.
    \item \textbf{Emphasis}: A distinction can be emphasized if it provides effective support for a concern/issue that the other case lacks effective support for.
    \item \textbf{Downplay}: A distinction can be downplayed if it provides effective support for a concern/issue for which the other case has alternative effective support.
    \item \textbf{Significance}: A distinction is significant if it can be emphasized and cannot be downplayed.
\end{itemize}

\paragraph{Step-by-Step Analysis Process}
\begin{enumerate}
    \item For the target distinction, find all concerns/issues it has effective support for.
    \item Check if the distinction can be emphasized (i.e., the other case lacks support for any of those concerns/issues).
    \item Check if the distinction can be downplayed (i.e., the other case has alternative support for any of those concerns/issues).
    \item Apply the significance rule.
\end{enumerate}

\paragraph{One-shot Example} [...example...]

\paragraph{Output Format} Provide your analysis in a JSON object: `{"significance": "false"}`.

\paragraph{Input}

Analyze the argumentative roles of the distinction `F6(p)`. Provide your response in the specified JSON format.

Factor Hierarchy:
[...]

Current Case (C1):
[...]

Precedent Case (C2):
[...]

\end{tcolorbox}

\subsection{Task 3: Identify All Significant Distinctions}
\begin{tcolorbox}[enhanced, float=h, width=\columnwidth, title=Prompt for Task 3: Identify All Significant Distinctions, label=box:task3]
\small
\paragraph{Goal} Your task is to identify ALL significant distinctions between a current case (C1) and a precedent case (C2). Cases in this task are represented by factors. You will be given a hierarchy on the factors, concerns and issues in the Mermaid diagram, which is a structure where each node is a factor, concern or issue and edge(s) are directed paths between two nodes.

\paragraph{Key Definitions}
(Same definitions for Distinction, Factor Representation for Cases, Effective Support, Emphasis, Downplay, and Significance as in Task 2)

\paragraph{Step-by-Step Analysis Process}
\begin{enumerate}
    \item First, parse the hierarchy and identify all distinctions between C1 and C2.
    \item For each distinction, evaluate its significance:
    \begin{itemize}
        \item Find all concerns/issues the distinction effectively supports.
        \item Check for emphasis potential (does the other case lack support?).
        \item Check for downplay potential (does the other case have alternative support?).
        \item A distinction is significant if it can be emphasized and cannot be downplayed.
    \end{itemize}
    \item Collect all distinctions that are determined to be significant.
\end{enumerate}

\paragraph{One-shot Example} [...example...]

\paragraph{Output Format} Provide your analysis in a JSON object: `{"significant\_distinctions": ["F6(p)", "F19(d)"]}`.

\paragraph{Input}

Identify ALL significant distinctions between the current case (C1) and the precedent case (C2). Provide your response in the specified JSON format.

Factor Hierarchy:
[...]

Current Case (C1):
[...]

Precedent Case (C2):
[...]

\end{tcolorbox}

\section{Legal Knowledge Hierarchy of the CATO Framework}

\begin{figure}[t]
\centering
\includegraphics[width=0.35\textwidth]{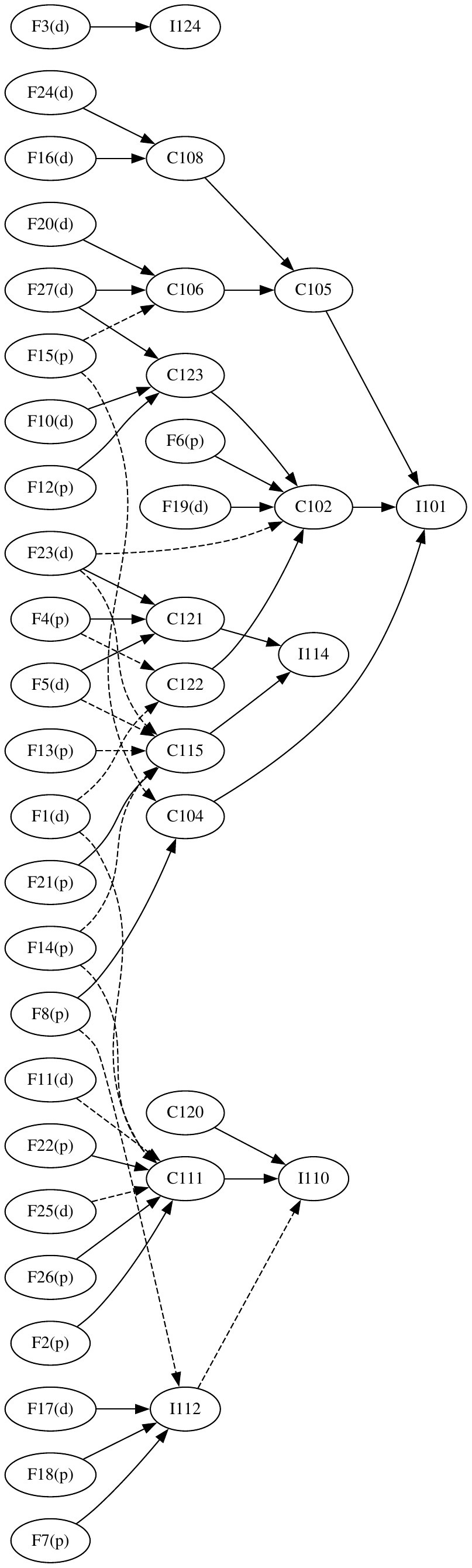}
\caption{Legal knowledge hierarchy in the CATO framework.}
\label{fig:cato}
\end{figure}

Figure \ref{fig:cato} illustrates the legal knowledge hierarchy used in our study.

\section{Rationale of Factors, Concerns and Issues in the CATO Framework}

Table \ref{tab:factormr}, Table \ref{tab:concerns}, and Table \ref{tab:issues} present the detailed rationales for the factors, concerns, and issues in the CATO framework.

\begin{table*}[h]
  \centering
  \scriptsize
  \caption{Legal Factors in the CATO Framework along with their meanings and rationales.}
  \label{tab:factormr}
  \begin{tabularx}{\linewidth}{p{3.5cm}XX}
  \toprule
  \textbf{Factor} & \textbf{Meaning} & \textbf{Rationale} \\
  \midrule
  F1 Disclosure-in-negotiations (D) & Plaintiff disclosed its product information in negotiations with defendant & Plaintiff gave his property away. \\
  F2 Bribe-employee (P) & Defendant paid plaintiff’s former employee to switch employment, apparently in an attempt to induce the employee to bring plaintiff’s information. & Defendant obtained plaintiff’s property through improper means. \\
  F3 Employee-sole-developer (D) & Employee defendant was the sole developer of plaintiff’s product. & Defendant should have property rights in his invention. \\
  F4 Agreed-not-to-disclose (P) & Defendant entered into a nondisclosure agreement with plaintiff. & Plaintiff takes reasonable steps to protect his property. \\
  F5 Agreement-not-specific (D) & The nondisclosure agreement did not specify which information was to be treated as confidential. & Plaintiff did not specify in what he claims a property interest. \\
  F6 Security-measures (P) & Plaintiff adopted security measures. & Plaintiff takes reasonable steps to protect his property. \\
  F7 Brought-tools (P) & Plaintiff’s former employee brought product development information to defendant. & Defendant steals plaintiff’s property. \\
  F8 Competitive-advantage (P) & Defendant’s access to plaintiff’s product information saved it time or expense. & Plaintiff’s trade secret is valuable property. \\
  F10 Secrets-disclosed-outsiders (D) & Plaintiff disclosed its product information to outsiders. & Plaintiff gave his property away. \\
  F11 Vertical-knowledge (D) & Plaintiff’s information is about customers and suppliers (which means that it may be available independently from customers or even in directories). & Plaintiff cannot have a property interest in its customer’s business info. \\
  F12 Outsider-disclosures-restricted (P) & Plaintiff’s disclosures to outsiders were subject to confidentiality restrictions. & Plaintiff protects his property. \\
  F13 Non-competition-agreement (P) & Plaintiff and defendant entered into a non-competition agreement. & Plaintiff protected against former employee’s use of confidential information. \\
  F14 Restricted-materials-used (P) & Defendant used materials that were subject to confidentiality restrictions. & Defendant used plaintiff’s property despite plaintiff’s protections. \\
  F15 Unique-product (P) & Plaintiff was the only manufacturer making the product. & Plaintiff’s trade secret is valuable property. \\
  F16 Info-reverse-engineerable (D) & Plaintiff’s product information could be learned by reverse-engineering. & Plaintiff’s property interest is limited in time. \\
  F17 Info-independently-generated (D) & Defendant developed its product by independent research. & Plaintiff has no property interest in information defendant generated independently. \\
  F18 Identical-products (P) & Defendant’s product was identical to plaintiff’s. & Defendant copied plaintiff’s trade secret property. \\
  F19 No-security-measures (D) & Plaintiff did not adopt any security measures. & Plaintiff did not protect his property. \\
  F20 Info-known-to-competitors (D) & Plaintiff’s information was known to competitors. & Plaintiff cannot have property interest in something known. \\
  F21 Knew-info-confidential (P) & Defendant knew that plaintiff’s information was confidential. & Defendant knew plaintiff claimed property interest. \\
  F22 Invasive-techniques (P) & Defendant used invasive techniques to gain access to plaintiff’s information. & Defendant used invasive techniques to steal plaintiff’s property. \\
  F23 Waiver-of-confidentiality (D) & Plaintiff entered into an agreement waiving confidentiality. & Plaintiff claimed no property interest in trade secret. \\
  F24 Info-obtainable-elsewhere (D) & The information could be obtained from publicly available sources. & Plaintiff cannot have property interest in something available from public sources. \\
  F25 Info-reverse-engineered (D) & Defendant discovered plaintiff’s information through reverse engineering. & Plaintiff’s property interest is limited by time. \\
  F26 Deception (P) & Defendant obtained plaintiff’s information through deception. & Plaintiff was cheated of his property \\
  F27 Disclosure-in-public-forum (D) & Plaintiff disclosed its information in a public forum. & Plaintiff gave his property interest in the trade secret away. \\
  \bottomrule
  \end{tabularx}
\end{table*}

\begin{table*}[h]
  \centering
  \scriptsize
  \caption{Intermediate Legal Concerns in the CATO Framework along with their meanings and rationales.}
  \label{tab:concerns}
  \begin{tabularx}{\linewidth}{p{3.5cm}XX}
  \toprule
  \textbf{Concern} & \textbf{Meaning} & \textbf{Rationale} \\
  \midrule
  C102 Efforts-To-Maintain-Secrecy (p) & Plaintiff took reasonable efforts to maintain the secrecy of the information. & Secrecy is a prerequisite for trade secret protection; reasonable efforts are required. \\
  C104 Info-Valuable (p) & The information has value to the plaintiff's business. & Information must have economic value to be protected as a trade secret. \\
  C105 Info-Known-Or-Available (d) & The information is known or available to people outside the plaintiff's business. & Information that is generally known or readily ascertainable is not a secret. \\
  C106 Info-Known (d) & The information is generally known in the relevant trade or industry. & Matters of general knowledge in an industry cannot be trade secrets. \\
  C108 Info-Available-Elsewhere (d) & The information is available from sources other than the plaintiff. & If information is public or available elsewhere, it lacks the necessary secrecy. \\
  C111 Questionable-Means (p) & Defendant used means that are questionable or unethical to acquire the information. & Using improper or unethical methods to acquire information supports misappropriation. \\
  C115 Notice-Of-Confidentiality (p) & Defendant had notice that the information was considered confidential. & Notice creates a duty to maintain confidentiality or abstain from use. \\
  C120 Info-Legitimately-Obtained-Or-Obtainable (d) & Defendant obtained the information legitimately or could have done so. & Legitimate acquisition (e.g., independent development) negates improper means. \\
  C121 Express-Confidentiality-Agreement (p) & An express confidentiality agreement existed between the parties. & Contractual agreements explicitly establish the duty of confidentiality. \\
  C122 Efforts-To-Maintain-Secrecy-Vis-A-Vis-Defendant (p) & Plaintiff took specific steps to maintain secrecy with respect to the defendant. & Specific precautions demonstrate the plaintiff's intent to keep info secret from defendant. \\
  C123 Efforts-To-Maintain-Secrecy-Vis-A-Vis-Outsiders (p) & Plaintiff took measures to maintain secrecy with respect to third parties. & General security measures are evidence of the information's secret status. \\
  \bottomrule
  \end{tabularx}
\end{table*}

\begin{table*}[h]
  \centering
  \scriptsize
  \caption{Legal Issues in the CATO Framework along with their meanings and rationales.}
  \label{tab:issues}
  \begin{tabularx}{\linewidth}{p{3.5cm}XX}
  \toprule
  \textbf{Issue} & \textbf{Meaning} & \textbf{Rationale} \\
  \midrule
  I101 Info-Trade-Secret (p) & Plaintiff's information constitutes a trade secret. & To establish a claim, the information must be a valid trade secret. \\
  I110 Improper-Means-Conclusion (p) & Defendant obtained the information through improper means. & Misappropriation by improper means is a violation of trade secret law. \\
  I112 Info-Used (p) & Defendant used the plaintiff's information. & Liability typically requires use or disclosure of the trade secret. \\
  I114 Confidential-Relationship (p) & A confidential relationship existed between plaintiff and defendant. & Misappropriation can be based on a breach of a confidential relationship. \\
  I124 Defendant-Ownership-Rights (d) & Defendant has ownership rights to the information. & If defendant owns the information, they cannot be liable for misappropriation. \\
  \bottomrule
  \end{tabularx}
\end{table*}

\end{document}